\PassOptionsToPackage{table}{xcolor}
\documentclass[11pt]{article}
\usepackage{float}
\usepackage{acl}
\usepackage{siunitx}
\usepackage[table]{xcolor}
\usepackage{graphicx}
\graphicspath{{./figures/}}
\usepackage{iftex}
\usepackage{fontspec}
\ifLuaTeX
  \usepackage[english,bidi=basic]{babel}
  \babelprovide[import]{arabic}

  \babelfont{rm}{Tinos}
  \babelfont[*arabic]{rm}[Scale=0.88]{Amiri}

  \providecommand{\LR}[1]{\foreignlanguage{english}{#1}}
\else
  \ifXeTeX
    \ExplSyntaxOn
    \cs_if_exist:NF \tbl_save_outer_table_cols:
      { \cs_new_protected:Npn \tbl_save_outer_table_cols: {} }
    \ExplSyntaxOff

    \usepackage{polyglossia}
    \setmainlanguage{english}
    \setotherlanguage{arabic}

    \newfontfamily\arabicfont[
      Script=Arabic,
      Scale=0.88,
      BoldFont=Amiri-Bold.ttf,
      ItalicFont=Amiri-Italic.ttf,
      BoldItalicFont=Amiri-BoldItalic.ttf
    ]{Amiri-Regular.ttf}
    \newfontfamily\arabicfontsf[
      Script=Arabic,
      Scale=0.88,
      BoldFont=Amiri-Bold.ttf
    ]{Amiri-Regular.ttf}
    \newfontfamily\arabicfonttt[
      Script=Arabic,
      Scale=0.88,
      BoldFont=Amiri-Bold.ttf
    ]{Amiri-Regular.ttf}

    \providecommand{\LR}[1]{\textenglish{#1}}
  \else
    \PackageError{HalluTruthQA}{Use LuaLaTeX or XeLaTeX}{%
      This document requires Unicode and Arabic font support.}
  \fi
\fi

\usepackage{enumitem}
\usepackage{changepage}
\usepackage[table,dvipsnames]{xcolor}
\usepackage{graphicx}
\usepackage{booktabs}
\usepackage{tabularx}
\usepackage{array}
\usepackage{multirow}
\usepackage{amsmath}
\usepackage{amssymb}
\usepackage{siunitx}
\usepackage{float}
\usepackage{verbatim}
\usepackage{xurl}
\usepackage{adjustbox}
\usepackage{needspace}
\usepackage[most]{tcolorbox}

\usepackage{tikz}
\usetikzlibrary{positioning,arrows.meta,calc,fit,backgrounds}

\usepackage{pgfplots}
\pgfplotsset{compat=1.18}

\usepackage{hyperref}
\newcolumntype{Y}{>{\raggedright\arraybackslash}X}

\colorlet{taxoHeader}{black!12}
\colorlet{taxoBlue}{blue!6}
\colorlet{taxoGreen}{green!7}
\colorlet{taxoOrange}{orange!10}
\colorlet{taxoPurple}{purple!7}
\colorlet{taxoRed}{red!7}

\definecolor{fcFill}{HTML}{EAF3FF}
\definecolor{fcLine}{HTML}{2F6FB3}

\definecolor{ciFill}{HTML}{EDF8EA}
\definecolor{ciLine}{HTML}{4C8A3F}

\definecolor{liFill}{HTML}{FFF1E8}
\definecolor{liLine}{HTML}{D8742F}

\definecolor{ffFill}{HTML}{F2ECFF}
\definecolor{ffLine}{HTML}{7A54B3}

\definecolor{nrFill}{HTML}{FFF0F0}
\definecolor{nrLine}{HTML}{C94C4C}

\definecolor{rootFill}{HTML}{F7F7F7}
\definecolor{rootLine}{HTML}{444444}
\definecolor{connGray}{HTML}{777777}

\definecolor{halluGray}{HTML}{BFBFBF}
\definecolor{halluFrame}{HTML}{D6D6D6}
\definecolor{halluRed}{HTML}{B00000}

\newtcolorbox{halluexamplebox}[1]{
  enhanced,
  colback=white,
  colframe=halluFrame,
  colbacktitle=halluGray,
  coltitle=white,
  boxrule=0.4pt,
  arc=2mm,
  left=6pt,
  right=6pt,
  top=5pt,
  bottom=5pt,
  title={#1},
  fonttitle=\bfseries,
  fontupper=\footnotesize,
  before skip=6pt,
  after skip=6pt
}

\newcommand{\HallBadge}[1]{%
\tikz[baseline=(B.base)]{%
\node[
    rounded corners=2pt,
    fill=white,
    draw=white,
    text=black!55,
    inner xsep=4pt,
    inner ysep=1pt,
    font=\sffamily\bfseries\scriptsize
] (B) {#1};%
}%
}

\newcommand{\HallTitle}[2]{%
\adjustbox{max width=.96\linewidth}{%
\HallBadge{#1}\hspace{0.45em}%
{\sffamily\bfseries\footnotesize #2}%
}%
}

\newenvironment{halluautoexample}[2]{%
  \Needspace{12\baselineskip}%
  \begin{halluexamplebox}{\HallTitle{#1}{#2}}%
  \footnotesize
}{%
  \end{halluexamplebox}%
}

\tikzset{
  exbadge/.style={
    circle,
    draw=black!45,
    fill=white,
    line width=0.45pt,
    font=\sffamily\bfseries\tiny,
    inner sep=0.8pt,
    minimum size=3.0mm
  }
}

\newcommand{\TaxMicroNode}[6]{%
\node[micro, draw=#4Line, fill=#4Fill] (#1) at (#2,#3) {#5};
\node[exbadge, anchor=north east, xshift=-1.6pt, yshift=-1.6pt]
at (#1.north east) {#6};
}

\newcommand{\halluArabicSize}{\fontsize{8pt}{9.3pt}\selectfont}

\newcommand{\halluEN}[1]{%
{\scriptsize\textit{#1}}\par\vspace{0.4mm}%
}

\newcommand{\halluError}[1]{%
{\scriptsize #1}%
}

\newcommand{\hallspan}[1]{%
\textcolor{red!70!black}{#1}%
}

\newcommand{\araboneline}[1]{%
\par\vspace{0.3mm}%
\begin{flushright}%
\begin{minipage}{0.88\linewidth}%
\begin{otherlanguage*}{arabic}%
\halluArabicSize
\raggedleft
#1%
\end{otherlanguage*}%
\end{minipage}%
\end{flushright}%
\vspace{-1.2mm}%
}

\newcommand{\arabjson}[1]{%
  \par\noindent
  \begin{minipage}{\linewidth}%
    \begin{otherlanguage*}{arabic}%
      \halluArabicSize
      \setlength{\parindent}{0pt}%
      \raggedleft
      #1\par
    \end{otherlanguage*}%
  \end{minipage}%
  \par\smallskip
}

\newcommand{\arabblock}[1]{%
\araboneline{#1}%
}

\newcommand{\arabpredoneline}[2]{%
\par\vspace{0.3mm}%
\begin{flushright}%
\begin{minipage}{0.88\linewidth}%
\begin{otherlanguage*}{arabic}%
\halluArabicSize
\raggedleft
{\color{black}#1}{\color{red!70!black}#2}%
\end{otherlanguage*}%
\end{minipage}%
\end{flushright}%
\vspace{-1.2mm}%
}

\newcommand{\halluAENline}[2]{%
\par\vspace{0.3mm}
\noindent
\begin{minipage}[t]{0.46\linewidth}
{\scriptsize\textit{#2}}
\end{minipage}
\hfill
\begin{minipage}[t]{0.50\linewidth}
\begin{otherlanguage*}{arabic}
\halluArabicSize
\raggedleft
#1
\end{otherlanguage*}
\end{minipage}
\par\vspace{0.4mm}
}

\newcommand{\ExRef}[2]{\hyperref[#1]{Example~#2}}

\usepackage{xcolor}
\usepackage{tikz}

\definecolor{affblue}{HTML}{24527A}

\newcommand{\authmark}[1]{%
  \raisebox{0.65ex}{%
    \tikz[baseline=(n.base)]
    \node[
      circle,
      fill=affblue,
      text=white,
      inner sep=0.65pt,
      font=\sffamily\tiny\bfseries
    ] (n) {#1};
  }%
}

\newcommand{\affmark}[1]{%
  \tikz[baseline=(n.base)]
  \node[
    circle,
    fill=affblue,
    text=white,
    inner sep=0.75pt,
    font=\sffamily\tiny\bfseries
  ] (n) {#1};
}

\usepackage{alphalph}

\title{HalluTruthQA: A Fine-Grained Benchmark for Hallucination Detection, Localization, and Explanation in Arabic QA}
\author{%
  \normalfont
  Abdessalam Bouchekif~\authmark{1},
  Mohammed-En-Nadhir Zighem~\authmark{}
  Salah Eddine Bekhouche~\authmark{3}\\[0.18em]
  Hichem Telli~\authmark{2}
  Somaya Eltanbouly~\authmark{1}
  Shahd Gaben~\authmark{1}
  Heba Sbahi~\authmark{1}
  Samer Rashwani~\authmark{1}\\[0.18em]
  Mutaz Al-Khatib~\authmark{1}
  Emad Mohamed~\authmark{4}
  Mohammed Ghaly~\authmark{1}
  Abdenour Hadid~\authmark{5}\\[1.05em]
  {\small
  \renewcommand{\arraystretch}{1.15}%
  \begin{tabular}{@{}c@{}c@{}}
    \affmark{1}\; Hamad Bin Khalifa University, Qatar
    &
    \affmark{2}\; University of Biskra, Algeria\\
    \affmark{3}\; University of the Basque Country, Spain
    &
    \affmark{4}\; Nazarbayev University, Kazakhstan\\
    \multicolumn{2}{c}{
      \affmark{5}\; Universiti Malaysia Kelantan, Malaysia
    }
  \end{tabular}}%
}

\begin{document}
\makeatletter
\patchcmd{\@maketitle}
  {\vskip 0.2in plus 1fil minus 0.1in}
  {\vskip 0.04in plus 0.1fil}
  {}{}
\makeatother
\maketitle

\begin{abstract}
Large language models (LLMs) can generate fluent Arabic answers, yet factual errors remain difficult to detect, localize, explain, and verify. Existing hallucination benchmarks often provide only response-level labels, offering limited support for identifying the exact erroneous content, explaining why it is incorrect, or recovering the correct factual answer. We introduce \textit{HalluTruthQA}, a fine-grained benchmark for hallucination evaluation in Arabic question answering. HalluTruthQA contains 2,400 expert-curated examples across four knowledge-intensive domains: Islamic knowledge, history, science, and geography. Each example includes an Arabic question, a model-generated answer, a verified reference answer, a binary hallucination label, and six candidate answers for factual verification. For hallucinated answers, the benchmark additionally provides character-level erroneous spans, human-written explanations, and macro- and micro-level hallucination types.
\\
We evaluate four open-source LLMs---ALLaM-7B, Falcon-H1R-7B, Qwen3-32B, and SILMA---in a zero-shot setting across four tasks: hallucination detection, span-level localization, factual verification, and explanation evaluation. The results show that these tasks measure complementary capabilities, with no single model achieving the best performance across all of them. The best observed scores are 0.880 Macro-F1 for hallucination detection, 0.516 F1-Sp for span localization, 0.852 LO-Score for factual verification, and 0.644 for explanation evaluation. Our findings highlight the importance of moving beyond binary hallucination detection toward fine-grained evaluation that also localizes, verifies, categorizes, and explains factual errors. The code, dataset, prompts, and evaluation scripts are available at
\url{https://gitlab.com/nlpresearcher/HalluTruthQA}.
\end{abstract}
\section{Introduction}

Large language models  have achieved strong performance in open-domain and knowledge-intensive question answering (QA). While they generate fluent and persuasive answers, their factual reliability remains a  challenge. LLMs are prone to hallucinations: statements that are unsupported, fabricated, or factually incorrect despite appearing plausible \cite{ji2023survey, huang2025survey}. These errors may involve wrong entities, dates, numerical values, locations, quotations, source attributions, or explanatory claims. In QA, even a short erroneous span can make an otherwise fluent answer unreliable.  Most hallucination evaluation resources rely on response-level labels indicating whether an answer is correct or hallucinated. While useful, such labels provide limited insight into the nature of the error. A response may contain the correct short answer but still include a fabricated source, an incorrect quotation, a wrong date, or an unsupported explanation. In these cases, binary evaluation fails to identify where the error occurs, why it is wrong, or whether the model can select the correct  answer. This motivates a fine-grained approach to hallucination evaluation that integrates answer-level detection, span-level localization, explanation, and verification. 
\\
This problem remains underexplored in Arabic. While existing Arabic resources have made progress, they do not provide comprehensive support for knowledge-intensive QA, exact character-level span annotations, human-written explanations, and candidate-based factual verification within a single benchmark. This gap is critical, as Arabic QA often involves domains where factual accuracy depends on fine-grained details and source-sensitive verification.
\\
Islamic knowledge, in particular, represents a sensitive setting. An answer may be superficially correct while grounding its claim in an incorrect Qur'anic reference, a fabricated hadith attribution, an invented quotation, or an unsupported legal or doctrinal claim. In this context, factual reliability depends not only on the final answer but also on the faithfulness of the supporting evidence. Similar challenges arise in history, science, and geography, where errors involve temporal reasoning, numerical values, spatial relations, and distinctions between closely related entities.

We introduce \textsc{HalluTruthQA}, a fine-grained benchmark for hallucination evaluation in Arabic QA. The benchmark contains 2,400 expert-curated QA examples across four knowledge-intensive domains: Islamic knowledge, history, science, and geography. Each example pairs an Arabic question and a model-generated answer with a verified reference answer, a binary hallucination label, six candidate answers for factual verification, and, for hallucinated answers, character-level erroneous spans accompanied by human-written explanations.
We evaluate four open-source LLMs in a zero-shot setting across the four tasks. Our results show that these tasks capture distinct challenges: models that perform well at detecting hallucinations do not necessarily localize erroneous spans, select the correct answer, or provide faithful explanations.
\\
This work makes three contributions:
\begin{itemize}[leftmargin=1.2em, labelsep=0.4em, itemsep=0pt, topsep=0pt, parsep=0pt]
\item We introduce \textsc{HalluTruthQA}, a fine-grained Arabic QA hallucination benchmark with 2,400 expert-curated QA examples across four knowledge-intensive domains.
\item We provide a multi-layer annotation scheme that includes response-level hallucination labels, character-level erroneous spans, human-written explanations, and six candidate answers for factual verification.
\item We evaluate four open-source LLMs in a zero-shot setting across hallucination detection, span-level localization, factual verification, and explanation evaluation.
\end{itemize}

\section{Related Work}
\label{sec:related_work}

Hallucination evaluation has become a central topic in large language
model research. Hallucinations are commonly divided into factuality
hallucinations, where generated claims contradict verifiable knowledge
or introduce unsupported information, and faithfulness hallucinations,
where the output is inconsistent with the input, context, instructions,
or its own reasoning \citep{huang2025survey}. Existing detection approaches include fact verification, uncertainty estimation, self-consistency, textual entailment, QA-based consistency checking, and LLM-based evaluation. They differ in how precisely they identify hallucinations, from assigning a label to the whole response to detecting specific erroneous parts of the generated text.

Several benchmarks study hallucination across QA,
summarization, retrieval-augmented generation, machine translation,
and multilingual generation. TruthfulQA \citep{lin2022truthfulqa} uses
adversarial questions designed to elicit common misconceptions, while
HaluEval \citep{li2023halueval} combines automatically generated and
human-annotated examples. FELM \citep{chen2023felm} provides
fine-grained factuality annotations across multiple domains, and
RAGTruth \citep{niu2024ragtruth} focuses on retrieval-augmented
generation. Mu-SHROOM \citep{vazquez2025mushroom} formulates
multilingual hallucination detection as span labeling, whereas
HalluVerse25 \citep{abdaljalil2025halluverse25} creates multilingual
examples by injecting controlled errors into factual biographical
sentences. HalOmi \citep{dale2023halomi} studies hallucination and
omission detection in multilingual machine translation, including
Arabic, while Poly-FEVER \citep{zhang2025polyfever} addresses
multilingual factual verification and hallucination detection across
multiple languages, including Arabic. These resources differ in
language coverage, task setting, grounding information, and annotation
granularity, but few jointly support detection, exact localization,
explanation, and factual verification. Arabic hallucination resources
remain comparatively limited.

\textsc{Halwasa} \citep{mubarak2024halwasa} focuses on isolated
LLM-generated sentences conditioned on predefined keywords rather than
natural QA. Multilingual benchmarks provide partial
Arabic coverage: HalluVerse25 includes Arabic in a controlled-error
setting, while Mu-SHROOM provides span-level annotations for Modern
Standard Arabic. Follow-up work has explored semantic-role
decomposition and textual entailment for Mu-SHROOM
\citep{elchafei2025hallucination}; however, these multilingual settings
are not specifically designed for knowledge-intensive Arabic QA. Other resources address Arabic generative QA more directly.

\textsc{IslamicEval} \citep{mubarak2025islamiceval} evaluates
hallucination detection and mitigation in Qur'anic and Hadith content,
while \textsc{Aftina} \citep{mohammed2025aftina} studies
retrieval-based mitigation for Islamic fatwa generation. Their
domain-specific scope, however, limits their coverage of broader Arabic
knowledge. \textsc{AraHalluEval}
\citep{alansari2025arahallueval} evaluates Arabic QA and summarization
using 12 factuality and faithfulness indicators, including entity and
numerical errors, contradictions, source conflicts, and fabrications.
\textsc{HalluScore} \citep{alansari2026halluscore} provides a
multi-domain Arabic QA benchmark covering adversarial formulations,
reasoning, historical knowledge, and Arabic cultural contexts, together
with verified evidence and reviewed answer explanations. Nevertheless,
neither resource defines exact character-level localization as a
general task for all hallucinated answers. In HalluScore, erroneous
spans are recorded only for partially hallucinated responses.

Fine-grained localization is important because response-level labels
cannot identify the precise nature or extent of an error. A response
may contain the correct main answer while adding an incorrect entity,
date, numerical value, quotation, attribution, or supporting claim.
Similarly, two responses with the same hallucination label may differ
substantially, with one being entirely incorrect and the other
containing only a short unsupported span. This limitation is especially
important in knowledge-intensive and source-sensitive domains. In
legal and religious QA, for example, an answer may appear persuasive
while misattributing a ruling, misquoting an authoritative text, or
fabricating supporting evidence. Prior work shows that LLMs can
misquote Qur'anic verses or generate unsupported evidence in Islamic
answers \citep{bouchekif-etal-2025-assessing}. Evaluation should
therefore assess not only overall correctness, but also source
attribution, evidential grounding, and the exact location of erroneous
content.

\textsc{HalluTruthQA} complements these resources by combining four
evaluation dimensions for knowledge-intensive Arabic question
answering: response-level hallucination detection, exact
character-level span localization, human-written explanations, and
candidate-based factual verification. Each model-generated answer is
checked against a verified reference answer, allowing the benchmark to
evaluate whether an answer is hallucinated, where the error occurs, why
it is incorrect, and whether the correct factual answer can be
selected. A discussion of hallucination causes and a feature-based
comparison with related Arabic and multilingual resources are provided
in Appendix~\ref{sec:def} and
Table~\ref{tab:hallucination_resources}.
\section{Data Description}

\begin{figure*}[t]
\centering
\includegraphics[width=0.95\textwidth,height=0.28\textheight]{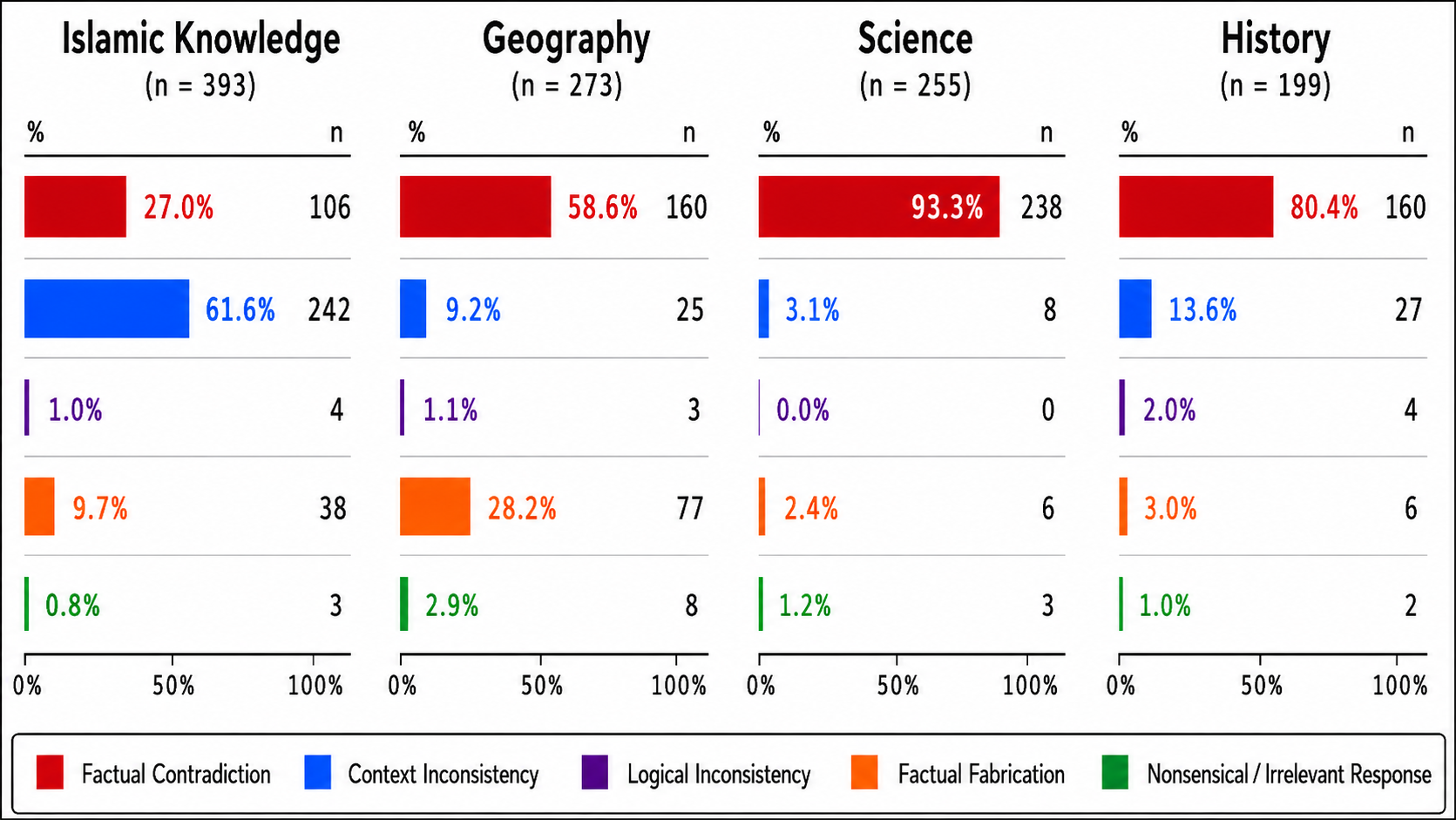}
\caption{Distribution of hallucination macro-types across domains.}
\label{fig:hallucination_type_distribution_by_domain}
\end{figure*}

\textsc{HalluTruthQA} contains 2,400 Arabic QA examples across four knowledge-intensive domains: Islamic knowledge, history, science, and geography (Table~\ref{fig:hallucination_type_distribution_by_domain}). Each example includes an Arabic question, a model-generated answer, a verified reference answer, and a binary hallucination label. For hallucinated answers, we provide character-level erroneous spans and human-written explanations. These spans are further categorized using a two-level hallucination
taxonomy. Macro-types capture failure categories, such as factual contradiction, context inconsistency, and factual fabrication, whereas micro-types identify the specific error affecting an annotated span, such as an incorrect entity, date, numerical value, citation, or source attribution. Each question is  paired with six manually constructed candidate answers: one  correct answer and five plausible distractors written in a style close to the generated output, so factual verification cannot rely on superficial stylistic cues. The benchmark supports hallucination detection, span-level localization, explanation evaluation, and multiple-choice factual verification.

\begin{table}[t]
\centering
\small
\setlength{\tabcolsep}{4.5pt}
\renewcommand{\arraystretch}{1.08}
\begin{tabular}{lccc}
\toprule
\textbf{Domain} & \textbf{Hallucinated} & \textbf{Non-hallucinated} & \textbf{Hallu. Spans} \\
\midrule
Islamic   & 341 (56.8\%)  & 259 (43.2\%)  & 393 \\
History   & 193 (32.2\%)  & 407 (67.8\%)  & 199 \\
Science   & 255 (42.5\%)  & 345 (57.5\%)  & 255 \\
Geography & 224 (37.3\%)  & 376 (62.7\%)  & 273 \\
\midrule
\rowcolor{gray!12}
\textbf{Total} & \textbf{1013 (42.2\%)} & \textbf{1387 (57.8\%)} & \textbf{1120} \\
\bottomrule
\end{tabular}
\caption{Distribution of hallucinated and non-hallucinated responses across domains. ``Hallucinated Spans'' denotes the total number of distinct text segments annotated as factually incorrect or unsupported within the generated responses. A single response may contain multiple hallucinated spans.}
\label{tab:dataset_distribution}
\end{table}

We use \textsc{Fanar-1-9B-Instruct}\footnote{\url{https://huggingface.co/QCRI/Fanar-1-9B-Instruct}} \citep{fanarteam2025fanar} as the single response generator for all questions. This provides a controlled and reproducible setting, where all answers are produced by the same Arabic-centric instruction-tuned model. It also avoids additional variation that could come from using several generators with different styles, strengths, and error patterns. Since the evaluated answers are generated by \textsc{FANAR-9B}, we additionally
report \textsc{FANAR-9B} as a self-detection reference setting to study whether a
model can detect, localize, verify, and explain hallucinations in its own
outputs.  These results are reported separately, rather than as evidence that \textsc{FANAR-9B} is superior to the other models. Answer generation used a maximum of 1024 new tokens, temperature 0.0, top-$p$ = 1.0, batch size 4, and \texttt{bfloat16} precision on a 48 GB GPU.

\subsection{Annotation Process}

Annotation followed two passes: expert annotation followed by verification
by trained research assistants. The four domains were handled by four
experts with relevant academic backgrounds, with one expert responsible
for each domain. This ensured that each subset was annotated by an expert
familiar with the corresponding knowledge area and its factual verification
needs. Annotators checked errors involving entities, dates, numerical
values, unsupported claims, fabricated references, and incorrect
attributions. For Islamic knowledge, they also verified Qur'anic references
and wording, hadith attributions, and unsupported religious claims. For hallucinated responses, annotators marked character-level erroneous
spans and wrote explanations. A response may contain multiple erroneous
spans. Each span is annotated independently as the minimal text segment
needed to isolate a single factual error. Each span is assigned one
macro-type and one micro-type; multi-label assignment is therefore not
used at the span level. When a response contains several distinct errors,
they are annotated as separate spans.
\\
Annotators manually created six candidate answers per question: one
verified correct answer and five plausible distractors. The distractors
were written in a style close to the generated response but were factually
incorrect, so factual verification could not rely on superficial stylistic
cues.

In the second pass, two research assistants reviewed all 2,400 examples. This pass flagged 84 examples (3.5\%), which were returned to the original domain experts for adjudication. Of these, 34 were confirmed without changes, 11 questions were revised for ambiguity, and 39 examples were revised, mainly to span boundaries. Before final adjudication, expert--reviewer agreement was high across annotation dimensions, with Cohen's $\kappa = 0.93$ for binary labels, $\kappa = 0.87$ for macro-types, $\kappa = 0.81$ for micro-types, 97.9\% agreement on the multiple-choice answer key, and span-level agreement of 0.89 character-F1 and 0.84 IoU. More annotation details are provided in Appendix~\ref{app:annotation_reliability}.

\subsection{Hallucination Type Distribution}

Figure~\ref{fig:hallucination_type_distribution_by_domain} shows the distribution of hallucination macro-types across domains. Across all hallucinated spans, \textit{Factual Contradiction} is the dominant category (59.3\%), typically involving wrong entities, dates, locations, events, numbers, roles, or concepts. \textit{Context Inconsistency} follows (27.0\%), covering source, attribution, or evidential mismatches. \textit{Factual Fabrication} accounts for 11.3\%, while \textit{Nonsensical / Irrelevant Response} and \textit{Logical Inconsistency} are rare (1.4\% and 1.0\%). At the domain level, \textit{Factual Contradiction} dominates in History, Science, and Geography, accounting for 80.4\%, 93.3\%, and 58.6\% of hallucinated spans, respectively. Islamic Knowledge differs: \textit{Context Inconsistency} accounts for 61.6\%, reflecting the need for faithful grounding in the appropriate verse, hadith, legal ruling, or source context. Geography shows the highest rate of Factual Fabrication (28.2\%).
Many geography questions require precise long-tail knowledge about
locations and spatial relations, which may lead models to generate
plausible but unsupported locations or overly specific geographic
explanations. Overall, hallucinations are usually fluent and relevant, but factually incorrect, unsupported or fabricated.
\\
These patterns highlight the need to evaluate attribution faithfulness and span-level localization beyond response-level detection. The complete hallucination taxonomy, including macro-type definitions, observed micro-types, is presented in Appendix~\ref{app:hallucination_taxonomy}, with   examples provided in Appendix~\ref{app:representative_hallucination_examples}.

\section{Experimental Setup}
\label{sec:experimental_setup}

\subsection{Evaluation Tasks}
We assess model performance across four tasks:

\begin{itemize}[leftmargin=1.2em, labelsep=0.9em, itemsep=3pt, topsep=0.8pt, parsep=0.4pt]
\item \textbf{Binary Hallucination Detection:} Given a question and a generated answer, the model predicts whether the response is \textsc{hallucinated} or \textsc{non-hallucinated}. This task evaluates the model's ability to detect factually incorrect, fabricated, or unsupported content.

\item \textbf{Span-level Localization:} For hallucinated instances, the model localizes erroneous spans at the character level. Performance is based on overlap between predicted and gold erroneous spans.

    \item \textbf{Explanation Evaluation:} The model generates an explanation identifying the erroneous content and justifying why it is factually incorrect. We use an LLM-as-a-judge protocol to compare generated explanations against human-written reference explanations. The judge assigns a score from 0 to 2 (0: incorrect/irrelevant; 1: partially correct; 2: complete/correct), normalized to $[0,1]$.

    \item \textbf{Multiple-Choice Factual Verification:} The model selects the correct option from six candidate answers, including one correct answer and five distractors. This evaluates whether the model can identify the correct factual answer among plausible alternatives.
\end{itemize}

\subsection{Models and Inference Protocol}

\paragraph{Models.} We evaluate four open-source LLMs: \textsc{ALLaM-7B}, \textsc{Falcon-H1R-7B}, \textsc{Qwen3-32B}, and \textsc{SILMA}. Since all evaluated answers are generated by \textsc{FANAR-9B}, we additionally report \textsc{FANAR-9B} as a self-detection reference setting to assess whether it can detect hallucinations in its own generated answers.

\paragraph{Inference Setup.} All models are evaluated in a zero-shot, closed-book setting without access to external knowledge bases or web search. To ensure a fair comparison, we use identical generation parameters: temperature $0.0$, top-$p=1.0$, and a maximum of 1024 new tokens. We use a two-stage prompting strategy: the first prompt asks the model to perform binary hallucination detection; for cases predicted as \textsc{hallucination}, the second prompt asks for span localization, explanation, and factual verification.

\begin{table*}[t]
\centering
\scriptsize
\setlength{\tabcolsep}{2.4pt}
\renewcommand{\arraystretch}{1.08}
\resizebox{\textwidth}{!}{%
\begin{tabular}{llcccccccccccc}
\toprule
\multirow{2}{*}{\textbf{Domain}} &
\multirow{2}{*}{\textbf{Model}} &
\multicolumn{4}{c}{\textbf{Hallucination Detection}} &
\multicolumn{3}{c}{\textbf{Span Localization}} &
\multicolumn{2}{c}{\textbf{MCQ Selection}} &
\multicolumn{3}{c}{\textbf{Explanation}} \\
\cmidrule(lr){3-6}
\cmidrule(lr){7-9}
\cmidrule(lr){10-11}
\cmidrule(lr){12-14}
& & \textbf{Acc.} & \textbf{Hall. F1} &
\textbf{No-Hall. F1} & \textbf{Macro-F1}
& \textbf{Span P.} & \textbf{Span R.} & \textbf{F1-Sp}
& \textbf{Correct} & \textbf{LO-Score}
& \textbf{Err. Ident.} & \textbf{Fact. Corr.} &
\textbf{Final Score} \\
\midrule

Islamic & \textsc{Qwen3-32B}
& 0.868 & 0.888 & 0.840 & \textbf{0.864}
& 0.850 & 0.344 & \textbf{0.489}
& 306/313 & 0.863 & 0.661 & 0.433 & 0.547 \\
& \textsc{Falcon-H1R-7B}
& 0.863 & 0.888 & 0.826 & 0.857
& 0.769 & 0.312 & 0.444
& 312/324 & 0.853 & 0.659 & 0.494 & 0.576 \\
& \textsc{ALLaM-7B}
& 0.862 & 0.876 & 0.844 & 0.860
& 0.839 & 0.248 & 0.383
& 247/293 & 0.823 & 0.668 & 0.530 & \textbf{0.599} \\
& \textsc{SILMA}
& 0.737 & 0.713 & 0.757 & 0.735
& 0.802 & 0.102 & 0.181
& 81/196 & 0.641 & 0.243 & 0.119 & 0.181 \\
\rowcolor{gray!10}
& \textsc{FANAR-9B}
& 0.908 & 0.924 & 0.885 & 0.904
& 0.823 & 0.319 & 0.460
& 303/333 & 0.883 & 0.627 & 0.354 & 0.490 \\

\midrule

Geography & \textsc{Qwen3-32B}
& 0.895 & 0.870 & 0.912 & \textbf{0.891}
& 0.635 & 0.409 & \textbf{0.498}
& 207/211 & 0.892 & 0.594 & 0.584 & \textbf{0.589} \\
& \textsc{Falcon-H1R-7B}
& 0.853 & 0.825 & 0.874 & 0.849
& 0.463 & 0.410 & 0.435
& 202/208 & 0.848 & 0.593 & 0.554 & 0.574 \\
& \textsc{ALLaM-7B}
& 0.877 & 0.832 & 0.903 & 0.867
& 0.500 & 0.254 & 0.337
& 112/183 & 0.818 & 0.522 & 0.520 & 0.521 \\
& \textsc{SILMA}
& 0.820 & 0.697 & 0.872 & 0.784
& 0.524 & 0.166 & 0.252
& 76/124 & 0.780 & 0.341 & 0.321 & 0.331 \\
\rowcolor{gray!10}
& \textsc{FANAR-9B}
& 0.877 & 0.835 & 0.902 & 0.868
& 0.510 & 0.254 & 0.340
& 175/187 & 0.867 & 0.517 & 0.499 & 0.508 \\

\midrule

Science & \textsc{ALLaM-7B}
& 0.915 & 0.908 & 0.921 & 0.915
& 0.766 & 0.344 & \textbf{0.475}
& 243/253 & 0.907 & 0.761 & 0.723 & \textbf{0.742} \\
& \textsc{Qwen3-32B}
& 0.778 & 0.793 & 0.762 & 0.777
& 0.570 & 0.388 & 0.462
& 254/254 & 0.778 & 0.688 & 0.485 & 0.587 \\
& \textsc{Falcon-H1R-7B}
& 0.885 & 0.880 & 0.889 & 0.885
& 0.663 & 0.344 & 0.453
& 254/254 & 0.885 & 0.675 & 0.510 & 0.592 \\
& \textsc{SILMA}
& 0.955 & 0.947 & 0.961 & \textbf{0.954}
& 0.878 & 0.207 & 0.335
& 150/241 & 0.879 & 0.503 & 0.390 & 0.446 \\
\rowcolor{gray!10}
& \textsc{FANAR-9B}
& 0.965 & 0.960 & 0.969 & 0.965
& 0.849 & 0.408 & 0.551
& 253/255 & 0.963 & 0.685 & 0.466 & 0.575 \\

\midrule

History & \textsc{Qwen3-32B}
& 0.855 & 0.812 & 0.881 & 0.847
& 0.545 & 0.660 & \textbf{0.597}
& 175/188 & 0.844 & 0.689 & 0.611 & 0.650 \\
& \textsc{ALLaM-7B}
& 0.873 & 0.827 & 0.899 & 0.863
& 0.557 & 0.594 & 0.575
& 120/182 & 0.821 & 0.711 & 0.694 & \textbf{0.702} \\
& \textsc{Falcon-H1R-7B}
& 0.836 & 0.794 & 0.864 & 0.829
& 0.513 & 0.634 & 0.567
& 172/189 & 0.822 & 0.687 & 0.623 & 0.655 \\
& \textsc{SILMA}
& 0.881 & 0.819 & 0.912 & \textbf{0.865}
& 0.484 & 0.362 & 0.415
& 97/161 & 0.828 & 0.263 & 0.246 & 0.255 \\
\rowcolor{gray!10}
& \textsc{FANAR-9B}
& 0.902 & 0.863 & 0.923 & 0.893
& 0.634 & 0.696 & 0.664
& 166/186 & 0.885 & 0.683 & 0.521 & 0.602 \\

\midrule

Global & \textsc{Qwen3-32B}
& 0.849 & 0.842 & 0.855 & 0.849
& 0.656 & 0.426 & \textbf{0.516}
& 942/966 & 0.844 & 0.659 & 0.514 & 0.587 \\
& \textsc{Falcon-H1R-7B}
& 0.859 & 0.853 & 0.866 & 0.859
& 0.618 & 0.400 & 0.486
& 940/975 & 0.852 & 0.654 & 0.536 & 0.595 \\
& \textsc{ALLaM-7B}
& 0.882 & 0.865 & 0.894 & \textbf{0.880}
& 0.688 & 0.333 & 0.449
& 722/911 & 0.842 & 0.673 & 0.614 & \textbf{0.644} \\
& \textsc{SILMA}
& 0.848 & 0.799 & 0.878 & 0.838
& 0.699 & 0.188 & 0.296
& 404/722 & 0.782 & 0.351 & 0.272 & 0.312 \\
\rowcolor{gray!10}
& \textsc{FANAR-9B}
& 0.913 & 0.902 & 0.922 & 0.912
& 0.727 & 0.390 & 0.508
& 897/961 & 0.900 & 0.632 & 0.444 & 0.538 \\

\bottomrule
\end{tabular}%
}

\caption{Bold indicates the best performance for each metric within each domain
and overall among the independent evaluation models; FANAR-9B is reported
separately as a self-detection reference.}
\label{tab:domain_global_results}
\end{table*}

\subsection{Evaluation Metrics}
\label{sec:evaluation_metrics}

We evaluate models on hallucination detection, span-level localization, factual verification, and explanation quality.

\begin{itemize}[leftmargin=1.2em, labelsep=0.9em, itemsep=3pt, topsep=0.8pt, parsep=0.4pt]
    \item \textbf{Hallucination Detection:} We report accuracy and F1-score for both hallucinated and non-hallucinated
classes. We use Macro-F1 as the primary detection metric, as it equally weights
both classes and is robust to label imbalance.

\item \textbf{Span-level Localization:}
We report partial-credit span-level F1 (\textit{F1-Sp}) following
\cite{sky2024androids}. Although hallucinated spans are stored as character offsets,
we compute \textit{F1-Sp} after mapping predicted and gold spans to token sets.
Recall is the average coverage of each gold span by the predicted spans, while
precision is the average coverage of each predicted span by the gold spans.
\textit{F1-Sp} is the harmonic mean of these two values. Detection-label
mismatches receive a span score of zero, while true-negative examples are
excluded. This metric rewards partial span overlap without requiring exact
boundary matches.

    \item \textbf{Multiple-Choice Factual Verification:} We report \textit{Correct}, the number of correctly selected options among true-positive hallucination predictions, and \textit{LO-Score} (Label--Option Score), which  evaluates hallucination detection and option selection. For each example, LO-Score gives 1.0 for a correct \textsc{no-hallucination} prediction, 0.5 for a correct \textsc{hallucination} prediction with an incorrect option, 1.0 for a correct \textsc{hallucination} prediction with the correct option, and 0.0 for  incorrect label prediction.

\item \textbf{Explanation Evaluation.}
We evaluate explanation quality using \textsc{GPT-5.5} as an LLM judge. The
judge is given the question, the model answer, the gold erroneous span, the
human-written reference explanation, and the model explanation. It  assigns a score reflecting how well the model justifies why the response is hallucinated.  Explanation quality is computed only when both the gold label and the model
prediction are hallucinated. The prompt is provided in
Appendix~\ref{app:explanation_prompt}. To check the reliability of the 
scores, two annotators reviewed a random sample of $25$  explanations per
domain. This review did not reveal systematic disagreement between the judge
scores and the human-written  explanations.

\end{itemize}
Because span localization, factual verification, and explanation generation are
performed only after a model predicts \textsc{hallucination}, these scores should
be interpreted as pipeline metrics. Errors in binary hallucination detection can
therefore propagate to the downstream tasks.

\subsection{Results Analysis}

Table~\ref{tab:domain_global_results} compares model performance across
hallucination detection, span localization, multiple-choice factual
verification, and explanation quality. Since all evaluated answers were
generated by \textsc{FANAR-9B}, its self-detection result may partly reflect cues
from its own writing style and the way it expresses uncertainty in generated
answers. We therefore report \textsc{FANAR-9B} separately as a reference setting,
rather than as direct evidence of general superiority over the other models.

Among the independent models, \textsc{ALLaM-7B} shows the strongest
detection performance, achieving the highest global Macro-F1 score
(0.880) with balanced performance across the two classes. \textsc{Falcon-H1R-7B}
and \textsc{Qwen3-32B} obtain slightly lower Macro-F1 scores, with 0.859 and
0.849, respectively. \textsc{SILMA} shows a different behavior: its lower
hallucination F1 compared to its non-hallucination F1 indicates that it is more
conservative and misses more hallucinated responses.

Span-level localization shows that detecting a hallucinated answer is not
sufficient: models must also identify the erroneous portion of the response.
Under the partial-credit F1-Sp metric, \textsc{Qwen3-32B} achieves the best global
localization performance, with an F1-Sp of 0.516, a precision of 0.656, and the
highest recall of 0.426. \textsc{FANAR-9B} follows closely with an F1-Sp of 0.508
and obtains the highest span precision (0.727), although its lower recall
(0.390) indicates that it localizes fewer gold erroneous spans. \textsc{Falcon-H1R-7B}
ranks third with an F1-Sp of 0.486, followed by \textsc{ALLaM-7B} with 0.449.
These results reveal a precision--recall trade-off: although \textsc{FANAR-9B} and
\textsc{ALLaM-7B} produce  precise spans, they leave a substantial
proportion of the annotated erroneous content uncovered. \textsc{SILMA} remains
the weakest localization model, with an F1-Sp of 0.296 and particularly low
recall (0.188), consistent with its conservative hallucination detection
behavior.

The low F1-Sp scores show that span localization is substantially harder than binary hallucination detection. The difficulty is not only deciding whether an answer is hallucinated, but identifying the exact part of the Arabic response that makes it wrong. Many errors involve short factual units, such as dates, numbers, entities, verse references, or source attributions, where small boundary differences can reduce overlap. In other cases, the correct short answer is followed by hallucinated support, so models may select only the answer entity or miss the false supporting claim. Arabic also introduces boundary variation because clitics, attached particles, and surface forms can make character-level spans harder to align exactly. As a result, models may detect the hallucination correctly but produce spans that are too short, too broad, or shifted from the erroneous span.

For MCQs, \textsc{Falcon-H1R-7B} performs best among the independent models, with a
global LO-Score of 0.852, closely followed by \textsc{Qwen3-32B} (0.844) and
\textsc{ALLaM-7B} (0.842). \textsc{Falcon-H1R-7B} is  effective at
selecting the correct candidate answer once an error is identified, selecting
the wrong option in only 3.6\% of true-positive cases.

In terms of explanation quality, \textsc{ALLaM-7B} outperforms all evaluated
systems, including \textsc{FANAR-9B}. It achieves the highest Error Identification
(0.673) and Factual Correction (0.614) scores, leading to the best Final Score
of 0.644. This highlights a  difference between detection and explanation:
while \textsc{FANAR-9B} excels at hallucination detection and candidate-option
selection and remains competitive in span localization, \textsc{ALLaM-7B} provides
stronger natural-language explanations with clearer factual corrections.
\vspace{-0.5em}
\paragraph{Domain-Specific Insights.}
The domain-level results show that hallucination evaluation presents different
challenges across knowledge areas. In the Islamic Knowledge domain, factual
correction and evidence grounding are difficult for all models, as errors often
involve scriptural wording, hadith attribution, transmission-related
information, or source-specific religious context. Although \textsc{FANAR-9B} and
\textsc{Qwen3-32B} achieve strong detection performance, their factual-correction
scores remain lower than in other domains. This suggests that detecting a
hallucination does not necessarily mean that a model can recover or justify the
correct religious evidence. For span localization in Islamic Knowledge,
\textsc{Qwen3-32B} obtains the highest partial-credit F1-Sp score (0.489),
followed by \textsc{FANAR-9B} (0.460), \textsc{Falcon-H1R-7B} (0.444), and
\textsc{ALLaM-7B} (0.383). The relatively high span precision of \textsc{Qwen3-32B}
(0.850), \textsc{ALLaM-7B} (0.839), and \textsc{FANAR-9B} (0.823), together with much
lower recall, indicates that models often identify accurate erroneous spans
when they make a prediction, but miss a large portion of the full annotated
error. Overall, models can localize some explicit fabricated references or
incorrect religious claims, but they  struggle to cover all erroneous
content and to produce  factual corrections.

In Geography, \textsc{Qwen3-32B} achieves the strongest overall performance,
leading in Macro-F1, F1-Sp, LO-Score, and explanation quality. Its F1-Sp score
reaches 0.498, outperforming \textsc{Falcon-H1R-7B} (0.435), \textsc{FANAR-9B}
(0.340), and \textsc{ALLaM-7B} (0.337). This suggests that \textsc{Qwen3-32B} is more
effective at localizing errors involving geographic locations and spatial
claims.

In Science, detection performance is generally high across models. Including
the self-detection reference setting, \textsc{FANAR-9B} achieves the highest
Macro-F1 score (0.965) and the highest F1-Sp score (0.551). Among the
independent evaluation models, \textsc{SILMA} achieves the strongest detection
performance with a Macro-F1 of 0.954, while \textsc{ALLaM-7B} obtains the best
span localization score with an F1-Sp of 0.475 and the best explanation Final
Score of 0.742. These results suggest that scientific hallucinations often
involve erroneous values, entities, formulas, or causal claims, which can be
localized more easily than implicit or context-dependent errors. However,
substantial portions of the annotated erroneous spans remain uncovered.

In History, \textsc{FANAR-9B} achieves the best localization performance in the
self-detection reference setting, with an F1-Sp score of 0.664.  \textsc{Qwen3-32B} obtains the highest F1-Sp score (0.597),
followed by \textsc{ALLaM-7B} (0.575) and \textsc{Falcon-H1R-7B} (0.567). The higher
recall of \textsc{FANAR-9B} (0.696) and \textsc{Qwen3-32B} (0.660) indicates that
these models capture a larger share of historically erroneous content. At the
same time, \textsc{ALLaM-7B} provides the strongest explanation quality, with the
highest Final Score of 0.702.

\paragraph{Error Analysis.}
To better understand the  types of errors, we analyze the 790 unique
instances where at least one model makes a wrong prediction. Since the same
instance can be misclassified by more than one model, these instances produce
1,556 model-level errors in total. We group the most frequent errors into four
main families:

\begin{enumerate}[leftmargin=*, nosep]
\item \textit{Temporal, Numeric, and Unit Normalization (39.0\%):}
These errors occur when models fail to recognize that two different
formats express the same fact. For example, a model may mark an answer as
hallucinated because it says ``40 AH'', while the reference answer says
``661 CE''. Although the formats are different, they refer to the same
historical period. This shows that models still struggle with converting
dates, numbers, units, and other equivalent factual expressions
(see \hyperlink{ex:quran_fabrication0}{Example~Non-Hallucinated},
\hyperlink{ex:quran_fabrication2}{Example~2},
\hyperlink{ex:quran_fabrication3}{Example~3}, and
\hyperlink{ex:quran_fabrication17}{Example~17}).

\item \textit{Source Attribution and Evidence Verification (22.2\%):}
These errors are especially common in the Islamic domain. They occur when
the model gives the correct main answer but supports it with an incorrect
source, such as a fabricated verse reference, a wrong citation, or an
inaccurate attribution. In these cases, models often focus on the fact that
the short answer is correct, but they fail to verify whether the supporting
evidence is also correct
(see \hyperlink{ex:quran_fabrication11}{Example~11} and
\hyperlink{ex:quran_fabrication15}{Example~15}).

\item \textit{Correct Short Answer with Hallucinated Support (17.1\%):}
These errors occur when the model gives the correct short answer but adds
a false or unsupported explanation. For example, the named entity or final
answer may match the reference answer, but the justification contains an
incorrect claim. Detection models often miss these cases because they rely
too much on the overlap with the correct answer and ignore the hallucinated
information in the explanation
(see \hyperlink{ex:quran_fabrication9}{Example~9} and
\hyperlink{ex:quran_fabrication10}{Example~10}).

\item \textit{Fully Incorrect Answer (11.3\%):}
These errors occur when the model's main answer is factually incorrect. In
some cases, the generated answer may be entirely different from, or directly
contradict, the correct answer. The error affects the core answer rather
than only the explanation or supporting evidence
(see \hyperlink{ex:quran_fabrication7}{Example~7},
\hyperlink{ex:quran_fabrication8}{Example~8}, and
\hyperlink{ex:quran_fabrication11}{Example~11}).

\end{enumerate}

Together, these four families cover 89.6\% of the model-level errors. The
remaining 10.4\% correspond to mixed or less frequent cases that combine several
patterns, such as partial factual mismatches, weak evidence checking,
over-specific additions, or imprecise span localization. We do not treat these
cases as a separate family because they do not form a single recurring error
pattern.

\paragraph{Common Failure Cases.}
A total of $39$  examples triggered an identical failure across all five
evaluated systems, including \textsc{FANAR-9B} in the self-detection reference setting. The vast majority of these shared errors (79.5\%) are false positives, where a
valid response was uniformly misclassified as hallucinated. Only
20.5\% correspond to missed true hallucinations. These shared failures are
concentrated in History (18 cases), followed by Geography (10), Islamic
 (8), and Science (3).

These findings emphasize that resolving hallucinations in advanced Arabic
models requires moving beyond factual recall; models must recognize semantic
paraphrases, validate fine-grained source attributions, handle temporal and
numerical conversions, isolate correct short answers from faulty supporting
explanations, and localize the exact erroneous spans. Additional annotation and
evaluation details are provided in Appendix~\ref{app:annotation_reliability}.

\section{Conclusion}

We introduced \textsc{HalluTruthQA}, a fine-grained, expert-curated benchmark for evaluating hallucinations in Arabic QA. Spanning Islamic knowledge, history, science, and geography, \textsc{HalluTruthQA} provides response-level hallucination labels, character-level erroneous spans, human-written explanations, and multiple-choice factual verification candidates. This multidimensional design shifts hallucination evaluation from binary correctness toward a granular understanding of where and why models fail, and whether they can recover the correct factual information. Our evaluation of four open-source instruction-tuned LLMs, together with the
\textsc{FANAR-9B} self-detection reference setting, shows that hallucination
detection, span localization, factual verification, and explanation quality
remain distinct, non-trivial challenges.
A key finding is that strong detection performance does not necessarily imply
high-quality explanations or accurate span localization: \textsc{FANAR-9B} performs
strongly in the self-detection reference setting, \textsc{Qwen3-32B} achieves the
best global span localization, and \textsc{ALLaM-7B} provides the most reliable
factual explanations. Moreover, our results reveal  domain-specific patterns: Islamic Knowledge is  affected by evidence-grounding and source-attribution errors, whereas history, science, and geography are mainly affected by factual contradictions and fabricated details. These findings underscore the need for Arabic hallucination benchmarks that move beyond binary detection to assess attribution faithfulness, temporal consistency, and reasoning errors. \textsc{HalluTruthQA} provides a step toward more reliable Arabic QA systems. Future work will expand the benchmark to additional domains, broader model coverage, and training settings aimed at improving factual reliability in Arabic language models.
\section*{Limitations}

This work has several limitations.
\begin{itemize}[leftmargin=*, nosep]

\item First, \textsc{HalluTruthQA} focuses on four knowledge-intensive domains:
Islamic knowledge, history, science, and geography. Although these domains cover
diverse factual phenomena, the benchmark does not capture all types of Arabic
QA, such as legal, medical, financial, or conversational
settings.

\item Second, the model-generated answers in the dataset are produced using a
single LLM, \textsc{FANAR-9B}. This design enables controlled annotation and
analysis, but it may limit the diversity of hallucination patterns compared to
datasets built from multiple generation models. Future work can extend the
benchmark with answers generated by a  range of Arabic and multilingual
LLMs.

\item Third, span-level hallucination annotation is  challenging. In
some cases, hallucinated content may involve implicit reasoning errors,
unsupported claims, or source attribution errors that are difficult to isolate
with exact character boundaries. To reduce this issue, we rely on expert
validation and use overlap-based span evaluation, but boundary variation remains
possible.

\item Fourth, explanation quality is evaluated using an LLM-as-a-judge protocol,
with additional human validation on a sample of judged explanations. While this
helps make the evaluation scalable, automatic judging may still introduce bias
or inconsistency. More extensive human evaluation would further strengthen the
reliability of explanation assessment.

\item Finally, we deliberately focus on open models and exclude proprietary
API-based systems such as Gemini or Claude from the main evaluation. This choice
is motivated by fairness and reproducibility. Our benchmark is designed as a
closed-book evaluation, where all models are tested under the same inference
conditions and without access to external search. In contrast, API-based systems
may rely on changing backend versions, retrieval components, web search, or
system-level grounding that are not always fully controllable by researchers.
Including such systems could make it harder to ensure that the evaluation
measures the model itself rather than external information access. This design
also supports the analysis and improvement of open Arabic language models, which
can be inspected, reproduced, and adapted by the research community.

\end{itemize}

\appendix
\section{Hallucination in Large Language Models and Related Resources}
\label{sec:def}

\paragraph{Definition.}
LLMs can generate fluent and convincing answers that are not necessarily factually correct. Hallucination generally refers to generated content that is inaccurate, fabricated, misleading, unsupported by reliable evidence, or inconsistent with the provided input or context
\citep{ji2023survey,huang-etal-2025-survey}. Hallucinations are commonly divided into two broad categories. A \emph{factual hallucination} occurs when a generated statement contradicts verifiable real-world knowledge, invents an entity or event, or incorrectly attributes a fact to a person, place, time, or source. A \emph{faithfulness hallucination} occurs when the generated content is not supported by the input or the provided context, even when the statement may be factually correct in isolation.

\paragraph{Causes of hallucination.}
Hallucination should not be attributed to a single cause. It can emerge
at different stages of the LLM development and generation pipeline
\citep{huang-etal-2025-survey,alansari2026survey}. One important cause
is incomplete knowledge coverage: the information required to answer a
question may be absent from, rare in, or insufficiently represented in
the model's training corpus. As a result, the model may rely on related
language patterns and generate a plausible but unsupported answer.
Hallucinations may also arise from incorrect, outdated, noisy, or
conflicting information in the training data, which can lead the model
to learn unreliable associations.

This issue is particularly relevant for Arabic, where high-quality,
diverse, and domain-specific data remain limited, especially in
specialized areas such as Islamic knowledge and legal reasoning.
Several Arabic-language resources have been introduced to evaluate LLMs
on Islamic knowledge, legal reasoning, and general culture
\citep{alwajih2025palmx,qias2025,
bouchekif2026mawarith, qias2026, gaben2026islamicturathbench, eltanbouly2026fiqhretriever}. However, many of these resources rely
mainly on multiple-choice questions, where models are evaluated by
selecting the correct option. This evaluation setting does not capture
how models formulate complete answers. Without evaluating full generated
responses, it is difficult to assess model behavior in depth,
particularly whether the model introduces hallucinated claims,
unsupported explanations, fabricated evidence, or incorrect source
attributions.

The next-token prediction objective may further contribute to
hallucination because it encourages the model to generate linguistically
likely continuations rather than explicitly verify the factual
correctness of each statement. Fine-tuning biases, alignment data,
prompt ambiguity, insufficient contextual grounding, and decoding
decisions may also increase the likelihood of hallucination. Therefore,
a model may hallucinate not only because it has not encountered the
required information, but also because it fails to retrieve,
distinguish, or correctly use knowledge that was present during
training.

\paragraph{Related Arabic and multilingual resources.}
Several recent studies have investigated hallucination in Arabic and
multilingual LLMs from complementary perspectives.
\textsc{AraHalluEval} introduces a multidimensional manual evaluation
framework for Arabic generative QA and summarization,
using fine-grained factuality and faithfulness indicators
\citep{alansari2025arahallueval}.
\textsc{HalluVerse-M$^3$} provides multilingual and multitask examples
constructed through controlled editing and human validation
\citep{abdaljalil2026halluverse}.
\textsc{HalluScore} focuses on hallucination-prone Arabic questions
across multiple domains, cultural settings, and reasoning requirements,
and provides fine-grained labels and explanations
\citep{alansari2026halluscore}.
In the Islamic domain, \textsc{Aftina} studies hallucination mitigation
in fatwa generation using retrieval-augmented generation and re-ranking
\citep{mohammed2025aftina}, while \textsc{IslamicEval} evaluates the
identification, validation, and correction of Qur'anic and Hadith
quotations \citep{mubarak2025islamiceval}.
\\
These resources differ in their languages, domains, annotation
procedures, and levels of supervision. Importantly,
\textsc{AraHalluEval} and \textsc{HalluScore} already go beyond binary
response-level labels by providing fine-grained annotations, and
\textsc{HalluScore} additionally provides explanations. However, neither
resource provides exact character-level localization of the
hallucinated content within generated responses. Span annotations are
available in \textsc{IslamicEval}, but they identify complete intended
Qur'anic and Hadith quotation spans rather than the specific erroneous
segments within the overall response. In contrast,
\textsc{HalluTruthQA} directly annotates the exact character-level spans
responsible for factual errors and associates them with fine-grained
hallucination types, human-written explanations, and verified factual
answers. Table~\ref{tab:hallucination_resources} summarizes these
differences.

\begin{table*}[t]
\centering
\caption{Feature-based comparison of \textsc{HalluTruthQA} with recent
Arabic and multilingual hallucination resources. ``Partial'' indicates
that the feature is available only for a subset of the data or is used
only for system-level evaluation.}
\label{tab:hallucination_resources}
\scriptsize
\setlength{\tabcolsep}{2.5pt}
\renewcommand{\arraystretch}{1.20}

\resizebox{\textwidth}{!}{
\begin{tabular}{
>{\raggedright\arraybackslash}p{1.70cm}
>{\raggedright\arraybackslash}p{1.55cm}
>{\raggedright\arraybackslash}p{2.25cm}
>{\raggedright\arraybackslash}p{1.65cm}
>{\raggedright\arraybackslash}p{2.05cm}
>{\raggedright\arraybackslash}p{2.35cm}
>{\raggedright\arraybackslash}p{2.65cm}
>{\raggedright\arraybackslash}p{2.65cm}
>{\raggedright\arraybackslash}p{2.70cm}
}
\toprule
\textbf{Resource} &
\textbf{Language(s)} &
\textbf{Domain / Task} &
\textbf{Outputs} &
\textbf{Manual annotation} &
\textbf{Span annotation} &
\textbf{Fine-grained labels} &
\textbf{Explanations} &
\textbf{Verified answer / Correction} \\
\midrule

\textsc{AraHalluEval}
\citep{alansari2025arahallueval}
&
Arabic
&
Generative QA and abstractive summarization
&
LLM outputs
&
Yes; two annotators and expert adjudication
&
No
&
12 factuality and faithfulness indicators
&
No error-specific explanations
&
Reference answer for QA or source document for summarization
\\

\addlinespace

\textsc{HalluVerse-M}\textsuperscript{3}
\citep{abdaljalil2026halluverse}
&
English, Arabic, Hindi, and Turkish
&
QA and dialogue summarization
&
Controlled LLM-edited outputs
&
Human validation
&
No
&
Entity-, relation-, and sentence-level hallucination types
&
No
&
Original ground-truth output
\\

\addlinespace

\textsc{HalluScore}
\citep{alansari2026halluscore}
&
Arabic
&
Multi-domain generative QA
&
LLM outputs
&
Yes; human annotation of model responses
&
Partial; hallucinated text is recorded only for partial-hallucination cases, not as a general character-offset localization task
&
Factual, faithfulness, and partial hallucination labels, together with multi-label question metadata
&
Ground-truth answer explanations generated by an LLM and manually reviewed
&
Verified ground-truth answer and supporting source link
\\

\addlinespace

\textsc{Aftina}
\citep{mohammed2025aftina}
&
Arabic
&
Islamic fatwa QA
&
LLM outputs + RAG
&
Partial; expert evaluation of 210 generated outputs
&
No
&
No corpus-level hallucination taxonomy
&
No
&
Dar Al-Ifta QA corpus and retrieved supporting content
\\

\addlinespace

\textsc{IslamicEval}
\citep{mubarak2025islamiceval}
&
Arabic
&
Qur'an and Hadith
&
LLM outputs
&
Yes; Islamic-studies experts
&
Yes; intended Qur'anic and Hadith quotation spans
&
Correct Qur'an, Incorrect Qur'an, Correct Hadith, and Incorrect Hadith
&
No
&
Canonical correction for incorrect quotations, or an indication that no valid correction exists
\\

\addlinespace

\textbf{\textsc{HalluTruthQA}}

\textbf{Arabic}
&
\textbf{Islamic knowledge, history, science, and geography}
&
\textbf{LLM outputs}
&
\textbf{Yes; domain-expert annotation followed by independent verification}
&
\textbf{Yes; exact character-level erroneous spans}
&
\textbf{Macro- and micro-level hallucination types}
&
\textbf{Human-written explanations of the localized errors}
&
\textbf{Verified reference answer and six candidate answers}
\\

\bottomrule
\end{tabular}
}
\end{table*}

Overall, the existing resources address complementary aspects of hallucination evaluation. \textsc{AraHalluEval} provides broad multidimensional evaluation across two Arabic generation tasks, while \textsc{HalluVerse-M$^3$} supports controlled comparisons across languages and tasks. \textsc{HalluScore} emphasizes hallucination-prone Arabic questions and provides verified evidence, answer explanations, and human response-level annotations. \textsc{Aftina} primarily evaluates hallucination mitigation through retrieval and re-ranking, whereas \textsc{IslamicEval} specializes in the accurate identification and correction of Qur'anic and Hadith quotations. In contrast, \textsc{HalluTruthQA} jointly supports response-level detection, exact erroneous-span localization, macro- and micro-level error classification, human-written explanation, and factual-answer verification in knowledge-intensive Arabic QA.
\subsection{Question Construction and Verification}
\label{sec:question-construction}

All questions were manually prepared by domain experts based on their
knowledge and experience. For each question, the corresponding expert
verified the reference answer using reliable  sources.
Experts were instructed to formulate clear, knowledge-intensive
questions while maintaining diversity in topics and difficulty levels.
Table~\ref{tab:question_sources} provides representative examples of the
sources consulted for each domain; these examples are not intended to
constitute an exhaustive list. For Islamic Knowledge, experts additionally checked Qur'anic wording and
references, hadith attributions, and the factual support of religious
claims. During verification, questions and reference answers were reviewed
for  accuracy, clarity, and ambiguity before inclusion in the
benchmark.

\begin{table*}[t]
\centering
\small
\begin{tabularx}{\textwidth}{@{}p{2.4cm}X@{}}
\toprule
\textbf{Domain} & \textbf{Representative Sources} \\
\midrule

Islamic Knowledge &
IslamWeb; IslamQA;
\textarabic{1213 سؤال وجواب في القرآن الكريم}
(\textit{1,213 Questions and Answers on the Holy Qur'an}) by Ahmad bin Ali Abu Islam;
\textarabic{200 سؤال وجواب في العقيدة الإسلامية}
(\textit{200 Questions and Answers on Islamic Creed}) by Hafiz bin Ahmad Al-Hakami;
\textarabic{السيرة النبوية في سؤال وجواب}
(\textit{The Prophetic Biography in Questions and Answers}) by Abu Abdullah Muhammad Ali Samak;
\textarabic{أكثر من 1500 سؤال وجواب في القرآن الكريم}
(\textit{More Than 1,500 Questions and Answers on the Holy Qur'an}) by Muhsin Hussein Al-Ghamdi;
and \textarabic{الموسوعة القرآنية}
(\textit{The Qur'anic Encyclopedia}). \\

\midrule
History &
\textarabic{أطلس تاريخ الإسلام}
(\textit{Atlas of Islamic History});
\textarabic{المفصل في تاريخ العرب قبل الإسلام}
(\textit{The Detailed History of the Arabs Before Islam}). \\

\midrule
Geography &
\textarabic{أطلس الوطن العربي}
(\textit{Atlas of the Arab World}) by Yahya Nabhan;
\textarabic{جغرافية العالم الإسلامي}
(\textit{Geography of the Islamic World}) by Mahmoud Taha Abu Al-Ala;
and \textarabic{الموسوعة العربية العالمية}
(\textit{The Global Arabic Encyclopedia}). \\

\midrule
Science &
\textarabic{موسوعة العلوم والتقانات}
(\textit{Encyclopedia of Science and Technology});
\textarabic{الموسوعة العلمية الميسرة}
(\textit{The Concise Scientific Encyclopedia}), edited and reviewed by Ahmad Shafiq Al-Khatib;
and \textarabic{الموسوعة العربية العالمية}
(\textit{The Global Arabic Encyclopedia}). \\

\bottomrule
\end{tabularx}
\caption{Representative sources consulted by domain experts during question
construction and reference-answer verification.}
\label{tab:question_sources}
\end{table*}

After the initial annotation, two trained research assistants reviewed the questions, reference answers, hallucination labels, localized spans, explanations, and answer options. Unclear or disputed cases were returned to the original domain expert for final verification. The annotation process was based on meaning rather than exact string matching. Therefore, semantically equivalent answers were accepted when they expressed the same fact. This also applied to different date formats. For example, corresponding Hijri and Gregorian dates were accepted when the experts judged them to refer to the same historical event. In the example provided in Appendix~A, ``40 AH'' and ``661 CE'' were treated as equivalent. Since both the questions and hallucination annotations were manually reviewed, differences between Hijri and Gregorian date formats were not automatically labeled as hallucinations.

\section{Data Format}
\label{app:data}

Each instance is stored in a structured JSON format. The format includes the question, generated response, reference answer, response-level hallucination label, span-level annotations, explanations, multiple-choice candidate answers, and the correct option.

\begin{tcolorbox}[
    colback=gray!3,
    colframe=gray!45,
    title=\textbf{Hallucinated Answer},
    fonttitle=\bfseries,
    boxrule=0.5pt,
    arc=2mm,
    left=2mm,
    right=2mm,
    top=1mm,
    bottom=1mm,
    breakable
]
\scriptsize

\noindent\texttt{\{}\\
\begin{adjustwidth}{1em}{0em}

\noindent\texttt{"id": "islamicHallucination\_8",}\\
\noindent\texttt{"question": } 
\arabjson{"ما البلد الذي ذُكر اسمه في القرآن الكريم؟",}

\noindent\texttt{"generated\_answer": } 
\arabjson{"بلد واحد فقط ذُكر اسمه في القرآن الكريم وهو مصر، وذلك في سورة يوسف الآية 47:
\textcolor{red!70!black}{وَقَالَ الْمَلِكُ ائْتُونِي بِهِ فَلَمَّا جَاءَهُ الرَّسُولُ قَالَ ارْجِعْ إِلَى رَبِّكَ فَسْأَلْهُ مَا لِذِي الثَّدْيَيْنِ إِنَّ قَدْ ظَلَمْتُ إِنَّ}",}

\noindent\texttt{"gold\_answer": } 
\arabjson{"البلد الذي ذُكر اسمه صراحة في القرآن الكريم هو مصر، ومن ذلك قوله تعالى: «وَقَالَ الَّذِي اشْتَرَاهُ مِن مِّصْرَ لِامْرَأَتِهِ أَكْرِمِي مَثْوَاهُ»، وقوله تعالى: «ادْخُلُوا مِصْرَ إِن شَاءَ اللَّهُ آمِنِينَ",}

\noindent\texttt{"generator\_model": "QCRI/Fanar-1-9B-Instruct",}\\
\noindent\texttt{"label": "hallucination",}\\
\noindent\texttt{"hallucinations": [}\\
\begin{adjustwidth}{1em}{0em}
\noindent\texttt{\{}
\begin{adjustwidth}{1em}{0em}
\noindent\texttt{"span\_start": 67,}\\
\noindent\texttt{"span\_end": 75,}\\
\noindent\texttt{"hallucinated\_span": } 
\arabjson{"\textcolor{red!70!black}{الآية 47}",}

\noindent\texttt{"explanation": } 
\arabjson{"الإحالة إلى الآية غير صحيحة؛ اسم مصر لم يرد في يوسف: 47، وإنما ورد في مواضع أخرى من السورة، منها يوسف: 21 ويوسف: 99."}
\end{adjustwidth}
\noindent\texttt{\},}

\noindent\texttt{\{}
\begin{adjustwidth}{1em}{0em}
\noindent\texttt{"span\_start": 77,}\\
\noindent\texttt{"span\_end": 226,}\\
\noindent\texttt{"hallucinated\_span": } 
\arabjson{"\textcolor{red!70!black}{وَقَالَ الْمَلِكُ ائْتُونِي بِهِ فَلَمَّا جَاءَهُ الرَّسُولُ قَالَ ارْجِعْ إِلَى رَبِّكَ فَسْأَلْهُ مَا لِذِي الثَّدْيَيْنِ إِنَّ قَدْ ظَلَمْتُ إِنَّ}",}

\noindent\texttt{"explanation": } 
\arabjson{"هذا نص مخترع منسوب إلى القرآن، وفيه ألفاظ غير قرآنية مثل الثديين؛ والآية الصحيحة في يوسف: 47 تتحدث عن الزراعة والسنين: قال تزرعون سبع سنين دأبا."}
\end{adjustwidth}
\noindent\texttt{\}}\\
\end{adjustwidth}

\noindent\texttt{],}\\
\noindent\texttt{"options": \{}\\

\begin{adjustwidth}{1em}{0em}
\noindent\texttt{"A": }
\arabjson{"- البلد الذي ذُكر اسمه صراحة في القرآن الكريم هو مصر، وقد ورد اسمها في سياق قصة يوسف عليه السلام.",}

\noindent\texttt{"B": }
\arabjson{"- البلد الذي ذُكر اسمه صراحة في القرآن الكريم هو فلسطين، وقد ورد ذكرها في سياق الأرض المقدسة.",}

\noindent\texttt{"C": }
\arabjson{"- البلد الذي ذُكر اسمه صراحة في القرآن الكريم هو العراق، وقد ارتبط بعدد من قصص الأنبياء والأمم السابقة.",}

\noindent\texttt{"D": }
\arabjson{"- البلد الذي ذُكر اسمه صراحة في القرآن الكريم هو الأردن، وقد ارتبط بأحداث تاريخية وقعت في بلاد الشام.",}

\noindent\texttt{"E": }
\arabjson{"- البلد الذي ذُكر اسمه صراحة في القرآن الكريم هو اليمن، وقد ارتبط بقصة سبأ وملكة سبأ.",}

\noindent\texttt{"F": }
\arabjson{"- البلد الذي ذُكر اسمه صراحة في القرآن الكريم هو السعودية، وقد ارتبط بذكر مكة المكرمة والبيت الحرام."}
\end{adjustwidth}

\noindent\texttt{\},}\\
\noindent\texttt{"answer": "A"}

\end{adjustwidth}
\noindent\texttt{\}}

\end{tcolorbox}

For non-hallucinated responses, the span list is empty. This unified format allows the same dataset to be used for response-level hallucination detection, explainable hallucination analysis, span-level hallucination localization, multiple-choice factual verification, and training models to select or generate factually reliable answers.

\hypertarget{ex:quran_fabrication0}{}
\begin{tcolorbox}[
    colback=gray!3,
    colframe=gray!45,
    title=\textbf{\textbf{Non-Hallucinated Answer}},
    fonttitle=\bfseries,
    boxrule=0.5pt,
    arc=2mm,
    left=2mm,
    right=2mm,
    top=1mm,
    bottom=1mm,
    breakable
]
\scriptsize

\noindent\texttt{\{}
\begin{adjustwidth}{1em}{0em}

\noindent\texttt{"id": "historicHallucination\_474",}\\
\noindent\texttt{"question": } 
\arabjson{"في أي عام انتهى عصر الخلفاء الراشدين؟",}

\noindent\texttt{"generated\_answer": } 
\arabjson{"انتهى عصر الخلفاء الراشدين في العام 40 هـ.",}

\noindent\texttt{"gold\_answer": } 
\arabjson{"661 م",}

\noindent\texttt{"generator\_model": "QCRI/Fanar-1-9B-Instruct",}\\
\noindent\texttt{"label": "no\_hallucination",}\\

\end{adjustwidth}
\noindent\texttt{\}}

\end{tcolorbox}

\section{Annotation Process and Reliability}
\label{app:annotation_reliability}

\begin{figure*}[t]
    \centering
    \includegraphics[width=0.92\textwidth]{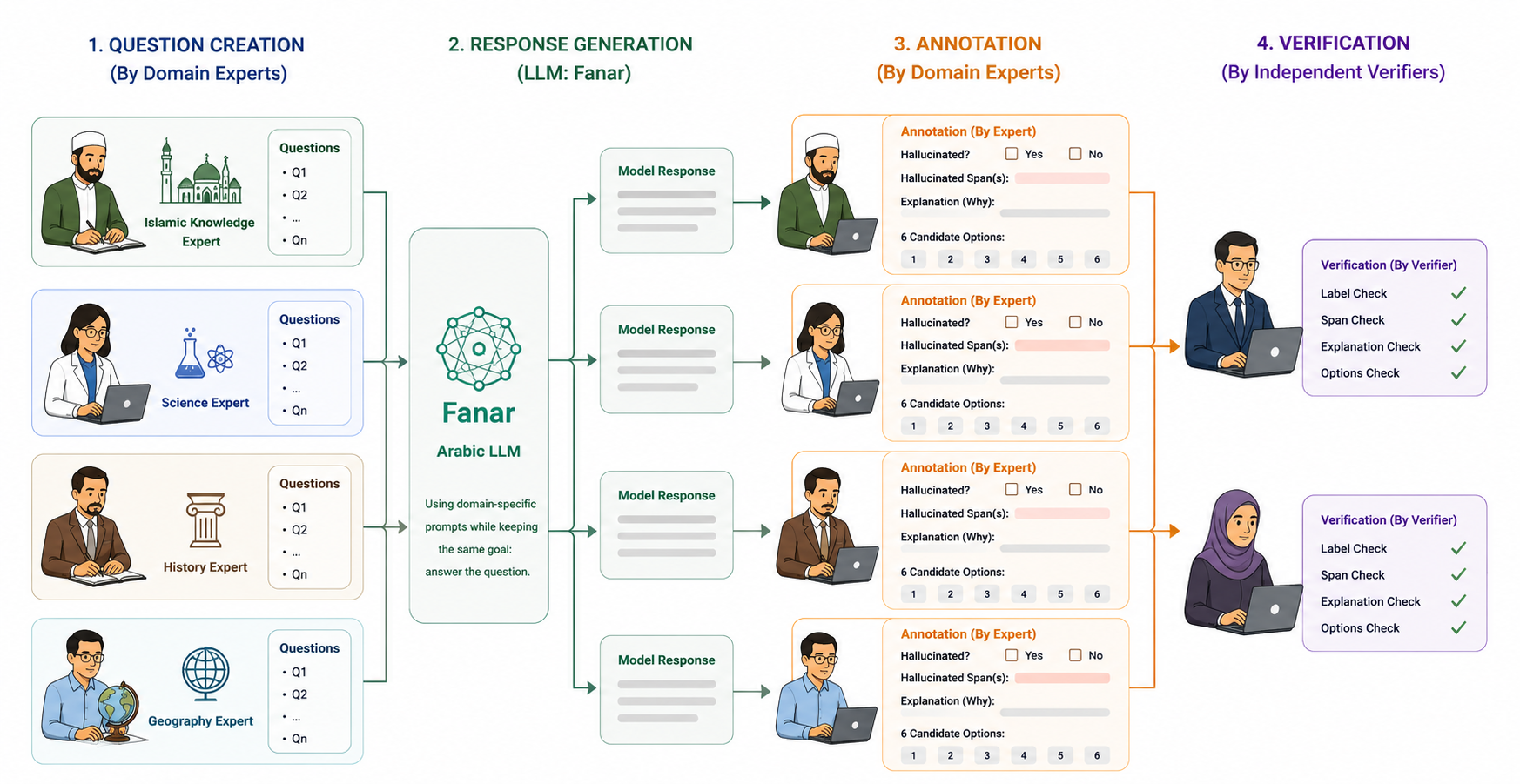}
    \caption{Overview of the \textsc{HalluTruthQA} construction pipeline.}
    \label{fig:hallutruthqa_construction}
\end{figure*}

The annotation process involved two stages. First, four domain experts annotated
the dataset, with each expert responsible for one domain: Islamic knowledge,
history, science, or geography. Each domain expert had a relevant academic
background in the corresponding knowledge area.
\\
The experts manually annotated the binary hallucination label, hallucination
macro-type, hallucination micro-type, erroneous spans, explanations, and
reference answers. They also manually constructed and validated the
multiple-choice candidate options and the corresponding answer key. Annotators
were instructed to ensure that the candidate options followed the same style,
wording, and level of specificity as the model-generated answer, so that the
correct option could not be identified from stylistic cues.

Second, two trained research assistants independently verified the annotations
produced by the four domain experts. The verification pass checked the binary
hallucination label, hallucination categories, span boundaries, explanations, and
the correctness of the multiple-choice answer key. Since the four domain experts
annotated disjoint domain-specific subsets, agreement was not computed among the
four experts on shared items. Instead, we report expert--reviewer agreement
between the initial domain-expert annotations and the independent verification
pass.

Disagreements or flagged cases were resolved through adjudication by the original
domain expert after the second verification pass.
Agreement was computed before final expert adjudication. We used Cohen's
$\kappa$ for categorical annotations and character-level F1 and
intersection-over-union (IoU) for span-level annotations. Disagreements and
flagged cases were subsequently resolved through adjudication by the original
domain expert after the second verification pass.

\begin{table}[!h]
\centering
\small
\setlength{\tabcolsep}{4pt}
\renewcommand{\arraystretch}{1.12}
\begin{tabular}{@{}l l c@{}}
\toprule
\textbf{Annotation dimension} & \textbf{Metric} & \textbf{Score} \\
\midrule
Binary hallucination label & Cohen's $\kappa$ & 0.93 \\
Macro hallucination type & Cohen's $\kappa$ & 0.87 \\
Micro hallucination type & Cohen's $\kappa$ & 0.81 \\
Multiple-choice answer key & Agreement & 0.979 \\
Span localization & Character-F1 & 0.89 \\
Span overlap & IoU & 0.84 \\
\bottomrule
\end{tabular}
\caption{Expert--reviewer agreement for HalluTruthQA before final adjudication.
Cohen's $\kappa$ is reported for categorical annotations, agreement is reported
for the multiple-choice answer key, and character-level F1 and IoU are used for
span-level annotations.}
\label{tab:annotation_reliability}
\end{table}

The agreement results indicate that the annotation process was reliable across
all dimensions. Binary hallucination labeling achieved the highest agreement
($\kappa=0.93$), showing that the domain-expert annotations and the independent
verification pass were highly consistent in deciding whether a generated answer
contained hallucinated content. Agreement remained strong for macro-level
hallucination categories ($\kappa=0.87$), suggesting that the main
hallucination types were clearly defined and consistently applied.
\\
Agreement was slightly lower for micro-level categories ($\kappa=0.81$), which
is expected given the difficulty of distinguishing closely related fine-grained
error types, such as unsupported claims, wrong attribution, citation mismatch,
and over-specific additions. Nevertheless, this score still indicates strong
agreement for fine-grained hallucination classification.
\\
The multiple-choice answer key achieved 97.9\% agreement, indicating that the
correct option was generally unambiguous. For span-level annotation, the
character-level F1 of 0.89 and IoU of 0.84 show that the reviewers largely
identified the same erroneous content as the original expert annotations.
Remaining differences were mainly due to span-boundary variation, such as
whether only the minimal erroneous entity, date, or reference was marked, or
whether surrounding explanatory context was also included.
\\
Overall, these results confirm that the annotations are robust. They also show
that response-level hallucination detection is the most stable annotation task,
whereas micro-type classification and exact span localization are more
challenging but still highly reliable.

%

\section{Hallucination Taxonomy}
\label{app:hallucination_taxonomy}

Figure~\ref{fig:hallucination_taxonomy_tree} presents the hierarchical
taxonomy of hallucination types used in \textsc{HalluTruthQA}.  The taxonomy was
designed to describe the erroneous spans observed during annotation. It contains
five macro-types: \textit{Factual Contradiction},
\textit{Context Inconsistency}, \textit{Logical Inconsistency},
\textit{Factual Fabrication}, and \textit{Nonsensical / Irrelevant Response}.
Each macro-type is further divided into fine-grained micro-types that describe
the specific form of the error.

\providecolor{rootLine}{HTML}{444444}
\providecolor{rootFill}{HTML}{F7F7F7}
\providecolor{connGray}{HTML}{888888}

\providecolor{fcLine}{HTML}{1E6ACB}
\providecolor{fcFill}{HTML}{EAF3FF}
\providecolor{ciLine}{HTML}{2E7D32}
\providecolor{ciFill}{HTML}{ECF8EC}
\providecolor{liLine}{HTML}{E86E1C}
\providecolor{liFill}{HTML}{FFF0E6}
\providecolor{ffLine}{HTML}{6F42C1}
\providecolor{ffFill}{HTML}{F3ECFF}
\providecolor{nrLine}{HTML}{D92D20}
\providecolor{nrFill}{HTML}{FFF1F0}

\providecommand{\TaxMicroNode}[6]{%
\node[micro#4] (#1) at (#2,#3) {#5};
\node[exbadge, anchor=north east, xshift=0.2mm, yshift=0.2mm] at (#1.north east) {#6};
}

\begin{figure*}[t]
\centering
\begin{adjustbox}{max width=\textwidth,center}
\begin{tikzpicture}[
    font=\sffamily,
    >=Latex
]

\tikzset{
    root/.style={
        rounded corners=5pt,
        draw=rootLine,
        fill=rootFill,
        line width=0.85pt,
        minimum width=3.8cm,
        minimum height=0.78cm,
        inner sep=3pt,
        align=center,
        font=\sffamily\bfseries\Large
    },
    macro/.style={
        rounded corners=4.5pt,
        line width=0.85pt,
        minimum width=2.70cm,
        minimum height=0.84cm,
        inner sep=2.5pt,
        align=center,
        font=\sffamily\bfseries\footnotesize
    },
    micro/.style={
        rounded corners=3.5pt,
        line width=0.72pt,
        minimum width=2.62cm,
        minimum height=0.62cm,
        text width=2.38cm,
        inner sep=2pt,
        align=center,
        font=\sffamily\bfseries\scriptsize
    },
    microfc/.style={micro, draw=fcLine, fill=fcFill},
    microci/.style={micro, draw=ciLine, fill=ciFill},
    microli/.style={micro, draw=liLine, fill=liFill},
    microff/.style={micro, draw=ffLine, fill=ffFill},
    micronr/.style={micro, draw=nrLine, fill=nrFill},
    exbadge/.style={
        circle,
        draw=black!35,
        fill=white,
        line width=0.35pt,
        inner sep=0.25pt,
        minimum size=2.7mm,
        font=\sffamily\bfseries\tiny
    },
    conn/.style={
        draw=connGray,
        line width=0.58pt
    }
}

\def\yRoot{4.05}
\def\yBar{3.22}
\def\yMacro{2.35}
\def\yTopMicro{1.08}
\def\stepY{0.78}

\def\xA{-6.90}
\def\xB{-3.45}
\def\xC{0.00}
\def\xD{3.45}
\def\xE{6.90}

\node[root] (root) at (0,\yRoot) {Hallucination};

\node[macro, draw=fcLine, fill=fcFill] (m1) at (\xA,\yMacro)
{1. Factual\\Contradiction};

\node[macro, draw=ciLine, fill=ciFill] (m2) at (\xB,\yMacro)
{2. Context\\Inconsistency};

\node[macro, draw=liLine, fill=liFill] (m3) at (\xC,\yMacro)
{3. Logical\\Inconsistency};

\node[macro, draw=ffLine, fill=ffFill] (m4) at (\xD,\yMacro)
{4. Factual\\Fabrication};

\node[macro, draw=nrLine, fill=nrFill] (m5) at (\xE,\yMacro)
{5. Nonsensical /\\Irrelevant Response};

\coordinate (hub) at (0,\yBar);
\coordinate (b1) at (\xA,\yBar);
\coordinate (b2) at (\xB,\yBar);
\coordinate (b3) at (\xC,\yBar);
\coordinate (b4) at (\xD,\yBar);
\coordinate (b5) at (\xE,\yBar);

\draw[conn] (root.south) -- (hub);
\draw[conn] (b1) -- (b5);

\draw[conn] (b1) -- (m1.north);
\draw[conn] (b2) -- (m2.north);
\draw[conn] (b3) -- (m3.north);
\draw[conn] (b4) -- (m4.north);
\draw[conn] (b5) -- (m5.north);

\TaxMicroNode{c11}{\xA}{\yTopMicro}{fc}{Entity Error}{1}
\TaxMicroNode{c12}{\xA}{\yTopMicro-\stepY}{fc}{Date / Temporal\\Error}{2}
\TaxMicroNode{c13}{\xA}{\yTopMicro-2*\stepY}{fc}{Numeric Error}{3}
\TaxMicroNode{c14}{\xA}{\yTopMicro-3*\stepY}{fc}{Location / Spatial\\Error}{4}
\TaxMicroNode{c15}{\xA}{\yTopMicro-4*\stepY}{fc}{Role / Status Error}{5}
\TaxMicroNode{c16}{\xA}{\yTopMicro-5*\stepY}{fc}{Event Error}{6}
\TaxMicroNode{c18}{\xA}{\yTopMicro-6*\stepY}{fc}{Causal Error}{7}
\TaxMicroNode{c17}{\xA}{\yTopMicro-7*\stepY}{fc}{Definition / Concept\\Error}{8}

\TaxMicroNode{c21}{\xB}{\yTopMicro}{ci}{Unsupported Claim}{9}
\TaxMicroNode{c22}{\xB}{\yTopMicro-\stepY}{ci}{Citation Mismatch}{10}
\TaxMicroNode{c23}{\xB}{\yTopMicro-2*\stepY}{ci}{Wrong Source\\Attribution}{11}
\TaxMicroNode{c24}{\xB}{\yTopMicro-3*\stepY}{ci}{Over-specific\\Addition}{12}

\TaxMicroNode{c31}{\xC}{\yTopMicro}{li}{Invalid Inference}{13}
\TaxMicroNode{c32}{\xC}{\yTopMicro-\stepY}{li}{Self-Contradiction}{14}
\TaxMicroNode{c33}{\xC}{\yTopMicro-2*\stepY}{li}{False Negation}{15}

\TaxMicroNode{c41}{\xD}{\yTopMicro}{ff}{Fabricated Entity}{16}
\TaxMicroNode{c42}{\xD}{\yTopMicro-\stepY}{ff}{Fabricated Source}{17}
\TaxMicroNode{c43}{\xD}{\yTopMicro-2*\stepY}{ff}{Fabricated Event}{18}
\TaxMicroNode{c44}{\xD}{\yTopMicro-3*\stepY}{ff}{Fabricated\\Explanation}{19}

\TaxMicroNode{c51}{\xE}{\yTopMicro}{nr}{Irrelevant Response}{20}
\TaxMicroNode{c52}{\xE}{\yTopMicro-\stepY}{nr}{Unintelligible Output}{21}
\TaxMicroNode{c53}{\xE}{\yTopMicro-2*\stepY}{nr}{Truncated Answer}{22}

\begin{pgfonlayer}{background}

\node[fit=(m1)(c17), rounded corners=5pt, fill=fcFill, draw=fcLine,
fill opacity=0.13, draw opacity=0.20, inner sep=3pt] {};

\node[fit=(m2)(c24), rounded corners=5pt, fill=ciFill, draw=ciLine,
fill opacity=0.13, draw opacity=0.20, inner sep=3pt] {};

\node[fit=(m3)(c33), rounded corners=5pt, fill=liFill, draw=liLine,
fill opacity=0.13, draw opacity=0.20, inner sep=3pt] {};

\node[fit=(m4)(c44), rounded corners=5pt, fill=ffFill, draw=ffLine,
fill opacity=0.13, draw opacity=0.20, inner sep=3pt] {};

\node[fit=(m5)(c53), rounded corners=5pt, fill=nrFill, draw=nrLine,
fill opacity=0.13, draw opacity=0.20, inner sep=3pt] {};

\end{pgfonlayer}

\end{tikzpicture}
\end{adjustbox}
\vspace{0.5mm}
\caption{Hierarchical taxonomy of hallucination macro-types and observed
micro-types.}
\label{fig:hallucination_taxonomy_tree}
\end{figure*}

\paragraph{Purpose of the taxonomy}
The goal of the taxonomy is to move beyond a binary distinction between
hallucinated and non-hallucinated answers. In Arabic QA, an
answer can be fluent and relevant but still contain a small factual error, a
wrong source attribution, an unsupported explanation, or a fabricated reference.
For this reason, each hallucinated answer in \textsc{HalluTruthQA} is annotated
not only with a response-level label, but also with character-level erroneous
spans, explanations, and hallucination types.
\paragraph{Macro and micro types}
Table~\ref{tab:macro_micro_definitions} defines the macro-types and their
corresponding micro-types. The macro-types capture the main ways in which a
model answer can fail, while the micro-types provide a more precise description
of the erroneous span. For example, a factual contradiction may involve an
entity, date, numeric, or definition error, whereas a context inconsistency may
involve a citation mismatch or a wrong source attribution.

\providecolor{fcLine}{HTML}{1E6ACB}
\providecolor{fcFill}{HTML}{EAF3FF}
\providecolor{ciLine}{HTML}{2E7D32}
\providecolor{ciFill}{HTML}{ECF8EC}
\providecolor{liLine}{HTML}{E86E1C}
\providecolor{liFill}{HTML}{FFF0E6}
\providecolor{ffLine}{HTML}{6F42C1}
\providecolor{ffFill}{HTML}{F3ECFF}
\providecolor{nrLine}{HTML}{D92D20}
\providecolor{nrFill}{HTML}{FFF1F0}

\begin{table*}[t]
\centering
\scriptsize
\setlength{\tabcolsep}{3.8pt}
\renewcommand{\arraystretch}{1.17}
\begin{tabularx}{\textwidth}{
@{}
>{\raggedright\arraybackslash}p{0.20\textwidth}
>{\raggedright\arraybackslash}p{0.17\textwidth}
>{\raggedright\arraybackslash}X
@{}}
\toprule
\textbf{Macro-type} & \textbf{Micro-type} & \textbf{Definition} \\
\midrule

\rowcolor{fcFill}
\multirow[t]{8}{0.19\textwidth}{
\textbf{\textcolor{fcLine}{Factual\\Contradiction}}\\[-0.2ex]
\textit{\textcolor{fcLine}{Wrong factual value}}
}
& Entity Error
& The response gives an incorrect person, country, organization, object, or named
entity instead of the verified answer. \\

\rowcolor{fcFill}
& Date / Temporal Error
& The response gives an incorrect date, year, period, duration, chronological
order, or temporal relation. \\

\rowcolor{fcFill}
& Numeric Error
& The response gives an incorrect number, quantity, percentage, distance, length,
count, or measurement. \\

\rowcolor{fcFill}
& Location / Spatial Error
& The response gives an incorrect place, region, country, spatial relation, or
geographical location. \\

\rowcolor{fcFill}
& Role / Status Error
& The response assigns an incorrect role, title, status, function, affiliation,
or relation to an entity. \\

\rowcolor{fcFill}
& Event Error
& The response describes a real or expected event incorrectly, assigns it to the
wrong actor, time, place, or outcome, or confuses it with another event. \\

\rowcolor{fcFill}
& Causal Error
& The response gives an incorrect cause, effect, motivation, or explanatory
relation between facts. \\

\rowcolor{fcFill}
& Definition / Concept Error
& The response confuses related concepts or gives a definition that contradicts
the verified answer. \\

\midrule

\rowcolor{ciFill}
\multirow[t]{4}{0.19\textwidth}{
\textbf{\textcolor{ciLine}{Context\\Inconsistency}}\\[-0.2ex]
\textit{\textcolor{ciLine}{Wrong evidence or grounding}}
}
& Unsupported Claim
& The response adds a claim that remains related to the question but is not
supported by the verified evidence. \\

\rowcolor{ciFill}
& Citation Mismatch
& The response links a claim to the wrong citation, verse, hadith, book, source,
or evidential reference. \\

\rowcolor{ciFill}
& Wrong Source Attribution
& The response attributes a statement to a source, scholar, text, or authority
that does not support that statement. \\

\rowcolor{ciFill}
& Over-specific Addition
& The response adds details that are more specific than what the verified
evidence allows, without necessarily inventing a new source or event. \\

\midrule

\rowcolor{liFill}
\multirow[t]{3}{0.19\textwidth}{
\textbf{\textcolor{liLine}{Logical\\Inconsistency}}\\[-0.2ex]
\textit{\textcolor{liLine}{Wrong reasoning}}
}
& Invalid Inference
& The response uses related information but draws a conclusion that does not
logically follow from it. \\

\rowcolor{liFill}
& Self-Contradiction
& The response contains two or more claims that are mutually incompatible within
the same answer. \\

\rowcolor{liFill}
& False Negation
& The response incorrectly denies a true relation or states the opposite of the
verified answer. \\

\midrule

\rowcolor{ffFill}
\multirow[t]{4}{0.19\textwidth}{
\textbf{\textcolor{ffLine}{Factual\\Fabrication}}\\[-0.2ex]
\textit{\textcolor{ffLine}{Invented factual content}}
}
& Fabricated Entity
& The response introduces a non-attested entity and presents it as factual. \\

\rowcolor{ffFill}
& Fabricated Source
& The response invents a source, citation, reference, verse, hadith, book, or
authority that does not exist or is not verified. \\

\rowcolor{ffFill}
& Fabricated Event
& The response describes an event that is not supported by the verified sources
and appears to be invented. \\

\rowcolor{ffFill}
& Fabricated Explanation
& The response gives an explanation that sounds specific or authoritative but is
not grounded in the verified evidence. \\

\midrule

\rowcolor{nrFill}
\multirow[t]{3}{0.19\textwidth}{
\textbf{\textcolor{nrLine}{Nonsensical /\\Irrelevant Response}}\\[-0.2ex]
\textit{\textcolor{nrLine}{No usable factual answer}}
}
& Irrelevant Response
& The response is off-topic or does not address the question in a meaningful way. \\

\rowcolor{nrFill}
& Unintelligible Output
& The response is too unclear, incoherent, or broken to be interpreted as a valid
factual answer. \\

\rowcolor{nrFill}
& Truncated Answer
& The response is incomplete or stops before providing a verifiable answer. \\

\bottomrule
\end{tabularx}
\caption{Macro-types, micro-types, and definitions in the
\textsc{HalluTruthQA} hallucination taxonomy. The macro-level cues clarify the
main distinction between wrong factual values, wrong grounding, wrong reasoning,
invented factual content, and unusable responses.}
\label{tab:macro_micro_definitions}
\end{table*}

\section{Hallucination Examples}
\label{app:representative_hallucination_examples}

To make the taxonomy easier to interpret, we provide examples for
the hallucination micro-types observed in \textsc{HalluTruthQA}. Each example is
labeled with its hallucination type and includes the question, the model
prediction, the verified answer, and a short explanation of the error.
\\
These examples illustrate how different hallucination types appear in Arabic
QA. Some cases involve direct factual contradictions, such as
wrong entities, dates, numbers, or locations. Other cases involve context
inconsistency, where the answer remains related to the question but the
supporting evidence, citation, or source attribution is incorrect. We also
include examples of logical inconsistency, factual fabrication, and nonsensical
or irrelevant responses.

\begin{halluautoexample}{1}{Factual Contradiction -- Entity Error}
\footnotesize

\textbf{Question}

\araboneline{من هو الصحابي الذي دعا له النبي صلى الله عليه وسلم بقوله: اللهم فقهه في الدين؟}

{\scriptsize\textit{Who was the Companion for whom the Prophet, peace be upon him, supplicated: ``O Allah, grant him understanding of the religion''?}}

\vspace{0.5mm}

\textbf{Prediction}

\araboneline{\textcolor{red!70!black}{معاذ بن جبل رضي الله عنه}}

{\scriptsize\textit{\textcolor{red!70!black}{Muadh ibn Jabal, may Allah be pleased with him.}}}

\vspace{0.5mm}

\textbf{Gold answer}

\halluAENline{الاسم هو عبد الله بن عباس}{Abdullah Ibn Abbas}

\textbf{Comment}\\
{\scriptsize The model predicts the wrong Companion instead of Abdullah Ibn Abbas.}
\end{halluautoexample}

\vspace{0.5mm}

\hypertarget{ex:quran_fabrication2}{}
\begin{halluautoexample}{2}{Factual Contradiction -- Date / Temporal Error}
\textbf{Question}

\araboneline{في أي عام تم توحيد ألمانيا؟}

\halluEN{In which year was Germany unified?}

\textbf{Prediction}

\araboneline{\hallspan{تم توحيد ألمانيا في العام 1871.}}

\halluEN{\hallspan{Germany was unified in the year 1871.}}

\textbf{Gold answer}

\araboneline{1990}

\textbf{Comment}

\halluEN{Although the German Empire was founded in 1871, the term ``German unification'' in contemporary contexts refers to the reunification of East and West Germany in 1990.}

\end{halluautoexample}

\hypertarget{ex:quran_fabrication3}{}
\begin{halluautoexample}{3}{Factual Contradiction -- Numeric Error}

\textbf{Question}

\araboneline{ما الطول التقريبي لنهر سانت لورانس، ومن أي بحيرة ينبع؟}

\halluEN{What is the approximate length of the Saint Lawrence River, and from which lake does it originate?}

\textbf{Prediction}

\araboneline{نهر سانت لورانس يمتد تقريبًا لـ \hallspan{3,058 كيلومتر (1,900 ميل)}. وهو ينبع من بحيرة أونتاريو، التي هي واحدة من البحيرات الكبرى في أمريكا الشمالية.}

\halluEN{The Saint Lawrence River extends for approximately \hallspan{3,058 kilometers (1,900 miles)}. It originates from Lake Ontario, which is one of the Great Lakes of North America.}

\textbf{Gold answer}

\araboneline{نحو 1,197 كم، وينبع من بحيرة أونتاريو.}

\halluEN{Approximately 1,197 km, and it originates from Lake Ontario.}

\textbf{Explanation}

\araboneline{الخطأ في الإجابة هو ذكر طول غير صحيح لنهر سانت لورانس. فطول النهر من بحيرة أونتاريو حتى المصب في خليج سانت لورانس نحو 1,197 كم، وليس 3,058 كم. وقد يشير الرقم 3,058 كم إلى منظومة البحيرات العظمى-سانت لورانس بأكملها، لا إلى النهر نفسه.}

\halluEN{The error is the incorrect length given for the Saint Lawrence River. The river’s length from Lake Ontario to its mouth in the Gulf of Saint Lawrence is approximately 1,197 km, not 3,058 km. The figure 3,058 km may refer to the entire Great Lakes--Saint Lawrence system, not to the river itself.}

\end{halluautoexample}

\begin{halluautoexample}{4}{Factual Contradiction -- Location / Spatial Error}

\textbf{Question}

\araboneline{ما المدينة التي تسمى 'مدينة الموتى' في مصر؟}

\halluEN{Which place is called the ``City of the Dead'' in Egypt?}

\textbf{Prediction}

\araboneline{المدينة التي تُعرف باسم ``مدينة الموتى'' في مصر هي \hallspan{مدينة الأقصر}. ولكن، من الجدير بالذكر أن هذا الاسم ليس رسمياً وقد يكون مستخدماً بشكل غير رسمي بسبب الكثافة العالية للمواقع الأثرية والمدافن القديمة فيها.}

\halluEN{The place known as the ``City of the Dead'' in Egypt is \hallspan{Luxor}. However, it should be noted that this name is not official and may be used informally because of the high density of archaeological sites and ancient tombs there.}

\textbf{Gold answer}

\araboneline{القرافة هي مدينة الموتى في قلب القاهرة الآن.}

\halluEN{Al-Qarafa is the City of the Dead, located today in the heart of Cairo.}

\textbf{Explanation}

\halluEN{The error is identifying Luxor as the ``City of the Dead'' in Egypt. This name refers to Al-Qarafa, the historic cemetery district in Cairo, not to the city of Luxor.}

\end{halluautoexample}

\hypertarget{ex:quran_fabrication5}{}
\begin{halluautoexample}{5}{Factual Contradiction -- Role / Status Error}

\textbf{Question}

\araboneline{من هو القائد الذي حرر الجزائر من الاستعمار الفرنسي سياسيًا؟}

\halluEN{Who was the leader who politically liberated Algeria from French colonialism?}

\textbf{Prediction}

\araboneline{القائد الذي حرر الجزائر من الاستعمار الفرنسي سياسيًا هو أحمد بن بلة. \hallspan{تولى رئاسة الحكومة المؤقتة للجمهورية الجزائرية في عام 1962 بعد مفاوضات إيفيان} التي أدت إلى استقلال البلاد عن فرنسا.}

\halluEN{The leader who politically liberated Algeria from French colonialism was Ahmed Ben Bella. \hallspan{He headed the Provisional Government of the Algerian Republic in 1962 after the Evian negotiations} that led to Algeria's independence from France.}

\textbf{Gold answer}

\halluAENline{يوسف بن خدة}{Benyoucef Benkhedda.}

\textbf{Comment}
\halluError{The model incorrectly attributes the presidency of the Provisional Government of the Algerian Republic during the Evian period to Ahmed Ben Bella. Ben Bella was imprisoned during the Evian negotiations, while Benyoucef Benkhedda headed the GPRA at that stage.}

\end{halluautoexample}


\begin{halluautoexample}{6}{Factual Contradiction -- Event Error}

\textbf{Question}

\araboneline{ما المعركة التي حدثت بين علي بن أبي طالب وبين الخوارج؟}

\halluEN{Which battle took place between Ali ibn Abi Talib and the Kharijites?}

\textbf{Prediction}

\araboneline{\hallspan{معركة الجمل هي المعركة التي حدثت بين علي بن أبي طالب والخوارج}. ولكن يجب التنويه إلى أن الخوارج لم يكونوا خصوم علي في هذه المعركة تحديداً، بل كانت \hallspan{معاوية بن أبي سفيان هو الخصم الرئيسي في معركة الجمل}، بينما كانت معركة النهروان هي المعركة التي دارت بين علي بن أبي طالب والخوارج. لكن إذا كنت تقصد أي معركة بين علي والخوارج، فإن معركة النهروان هي الأكثر شهرة.}

\halluEN{\hallspan{The Battle of the Camel was the battle that took place between Ali ibn Abi Talib and the Kharijites}. However, it should be noted that the Kharijites were not Ali's opponents in this specific battle; rather, \hallspan{Muawiya ibn Abi Sufyan was the main opponent in the Battle of the Camel}, whereas the Battle of Nahrawan was the battle that took place between Ali ibn Abi Talib and the Kharijites. But if you mean any battle between Ali and the Kharijites, then the Battle of Nahrawan is the most famous.}

\textbf{Gold answer}

\araboneline{المعركة التي حدثت بين علي بن أبي طالب والخوارج هي معركة النهروان عام 38 هـ.}

\halluEN{The battle that took place between Ali ibn Abi Talib and the Kharijites was the Battle of Nahrawan in 38 AH.}

\textbf{Explanation}

\halluEN{The first error is identifying the Battle of the Camel as the battle between Ali ibn Abi Talib and the Kharijites, whereas the correct battle is the Battle of Nahrawan. The statement that Muawiya ibn Abi Sufyan was the main opponent in the Battle of the Camel is also incorrect; Muawiya is mainly associated with the Battle of Siffin, not the Battle of the Camel.}

\end{halluautoexample}

\hypertarget{ex:quran_fabrication7}{}
\begin{halluautoexample}{7}{Factual Contradiction -- Causal Error}

\textbf{Question}

\araboneline{لماذا ينصح المرأة المرضع بعدم تناول الكحول؟}

\halluEN{Why is a breastfeeding woman advised not to consume alcohol?}

\textbf{Prediction}

\araboneline{\hallspan{يُسمح بالكحول أثناء الرضاعة بكميات معتدلة لأنه يُسرّع امتصاص الحديد في الحليب ولا يصل للرضيع.}}

\halluEN{\hallspan{Alcohol is allowed during breastfeeding in moderate amounts because it accelerates iron absorption in breast milk and does not reach the infant.}}

\textbf{Gold answer}

\araboneline{ينصح المرأة المرضع بعدم تناول الكحول لأن الكحول يمكن أن ينتقل إلى حليب الأم ويصل إلى الرضيع، وقد يؤثر في صحته ونموه ونمط نومه.}

\halluEN{A breastfeeding woman is advised not to consume alcohol because alcohol can pass into breast milk and reach the infant, potentially affecting the infant's health, development, and sleep patterns.}

\textbf{Explanation}

\halluEN{The error lies in giving a false causal justification for alcohol consumption during breastfeeding. Alcohol is not recommended because it does not accelerate iron absorption in breast milk, and the claim that it does not reach the infant is incorrect; alcohol can pass into breast milk and expose the infant.}

\end{halluautoexample}

\hypertarget{ex:quran_fabrication8}{}
\begin{halluautoexample}{8}{Factual Contradiction -- Definition / Concept Error}

\textbf{Question}

\araboneline{ما هي الميتة الحلال والدم الحلال؟}

\halluEN{What are the lawful dead animals and the lawful blood in Islamic law?}

\textbf{Prediction}

\araboneline{\hallspan{الميتة الحرام هي ما مات بغير ذكاة شرعية، أما الميتة الحلال فهي ما ماتت ذكاة شرعية. أما الدم الحرام فهو الدم الذي لا يجوز أكله، مثل دم الحيوانات الميتة أو التي لم تذبح بشكل صحيح، بينما الدم الحلال هو دم الحيوانات التي ذبحت بطريقة شرعية. ولكن يجب التنويه إلى أن الإسلام يحرم تناول الدم بشكل عام، سواء كان من حيوان مذبوح بشكل صحيح أم لا.}}

\halluEN{\hallspan{Forbidden carrion is what dies without lawful slaughter, whereas lawful carrion is what dies through lawful slaughter. Forbidden blood is blood that may not be consumed, such as the blood of dead animals or animals not slaughtered properly, whereas lawful blood is the blood of animals slaughtered according to Islamic law. However, it should be noted that Islam generally forbids consuming blood, whether from a properly slaughtered animal or not.}}

\textbf{Gold answer}

\araboneline{الميتة الحلال هي السمك والجراد، والدم الحلال هو الكبد والطحال.}

\halluEN{The lawful dead animals are fish and locusts, and the lawful blood refers to the liver and spleen.}

\textbf{Explanation}

\halluEN{The error lies in giving an incorrect definition of lawful carrion and lawful blood. Lawful carrion does not mean an animal that died through lawful slaughter; rather, it refers to the recognized exceptions: fish and locusts. Similarly, lawful blood does not mean the blood of a properly slaughtered animal, but refers to the liver and spleen.}

\end{halluautoexample}

\hypertarget{ex:quran_fabrication9}{}
\begin{halluautoexample}{9}{Context Inconsistency -- Unsupported Claim}

\textbf{Question}

\araboneline{ما هي أكبر جزيرة في العالم؟}

\halluEN{What is the largest island in the world?}

\textbf{Prediction}

\araboneline{جرينلاند هي الأكبر بمساحة تبلغ حوالي 2,166,086 كيلومتر مربع. ومع ذلك، يعتمد هذا على كيفية تعريف ``الجزر'' -- \hallspan{فجرينلاند تعتبر أحياناً شبه جزيرة بسبب اتصالها القاري مع كندا عبر مضيق دنمارك الصغير}.}

\halluEN{Greenland is the largest island, with an area of about 2,166,086 square kilometers. However, this depends on how ``islands'' are defined -- \hallspan{Greenland is sometimes considered a peninsula because of its continental connection with Canada through the narrow Denmark Strait}.}

\textbf{Gold answer}

\araboneline{غرينلاند}

\halluEN{Greenland.}

\textbf{Explanation}

\halluEN{The model gives the correct main answer, but then adds an unsupported claim about Greenland being sometimes considered a peninsula because of a supposed continental connection with Canada. This extra claim is not supported by the question or the reference answer and introduces unnecessary, ungrounded information.}

\end{halluautoexample}

\hypertarget{ex:quran_fabrication10}{}
\begin{halluautoexample}{10}{Factual Contradiction -- Location / Spatial Error}
\fontsize{8}{9}\selectfont

\textbf{Question}

\araboneline{ما اسم البحيرة الواقعة على حدود زيمبابوي وزامبيا والمعروفة بأنها من أكبر البحيرات الاصطناعية؟}

{\footnotesize\textit{What is the name of the lake located on the border between Zimbabwe and Zambia, known as one of the largest artificial lakes?}}

\vspace{1mm}

\textbf{Prediction}

\arabpredoneline
{البحيرة التي تشير إلى هي بحيرة كاريبا، وهي تقع على الحدود بين زيمبابوي وزامبيا، }
{وتشتهر بوجود شلالات فيكتوريا الشهيرة.}

{\footnotesize\textit{The lake referred to is Lake Kariba, located on the border between Zimbabwe and Zambia, and it is \textcolor{red!70!black}{famous for the presence of Victoria Falls}.}}

\vspace{1mm}

\textbf{Gold answer}

\araboneline{الاسم هو بحيرة كاريبا \LR{(Lake Kariba)}}

\vspace{1mm}

\textbf{Error.}
{\footnotesize The model identifies the correct lake, but adds a false geographical association with Victoria Falls.}

\end{halluautoexample}

\hypertarget{ex:quran_fabrication11}{}
\begin{halluautoexample}{11}{Factual Contradiction -- Citation Mismatch}

\textbf{Question}

\araboneline{هل سنّ النبي صلى الله عليه وسلم الإسراع بتجهيز الميت أم الإبطاء به؟}

\halluEN{Did the Prophet, peace be upon him, prescribe hastening the preparation of the deceased or delaying it?}

\textbf{Prediction}

\araboneline{\hallspan{لم يسن النبي صلى الله عليه وسلم الإسراع أو الإبطاء بشكل خاص بتجهيز الميت}، ولكن يُفضل الإسراع في تجهيزه دون تأخير طويل، \hallspan{وذلك لما روي عن عائشة رضي الله عنها أنها قالت: كان رسول الله صلى الله عليه وسلم إذا حضرته الوفاة جعل عنده شيئاً من التراب فيمسحه بوجهه ويقول: بِسْمِ اللَّهِ وَبِعِزَّةِ اللَّهِ، وَأَعُوذُ بِاللَّهِ  }.}

\halluEN{\hallspan{The Prophet, peace be upon him, did not specifically prescribe either hastening or delaying the preparation of the deceased}; however, it is preferable to prepare the deceased quickly without long delay, \hallspan{based on a report attributed to Aisha, may Allah be pleased with her, stating that when the Prophet was near death, he would place some dust near him, wipe his face with it, and say: In the name of Allah and by the might of Allah, and I seek refuge in Allah from...}.}

\textbf{Gold answer}

\araboneline{سنّ النبي صلى الله عليه وسلم الإسراع بتجهيز الميت ودفنه، وعدم تأخيره بلا حاجة.}

\halluEN{The Prophet, peace be upon him, prescribed hastening the preparation and burial of the deceased, and not delaying it without need.}

\textbf{Explanation}

\araboneline{الخطأ الأول هو نفي وجود سنة خاصة بالإسراع في تجهيز الميت، بينما السنة هي الإسراع بتجهيزه ودفنه وعدم تأخيره بلا حاجة. أما الخطأ الثاني فهو الاستدلال برواية غير مناسبة  لا تدعم الحكم المتعلق بتجهيز الميت.}

\halluEN{The first error is denying that there is a specific prophetic practice of hastening the preparation of the deceased, whereas the Sunnah is to prepare and bury the deceased promptly without unnecessary delay. The second error is the use of an inappropriate and incomplete narration that does not support the ruling about preparing the deceased.}

\end{halluautoexample}


\begin{halluautoexample}{12}{Context Inconsistency -- Wrong Source Attribution}

\textbf{Question}
\arabblock{أشار القرآن الكريم إلى عدة مهن كانت تزاول في الأمم السابقة، منها مهنة بناء السفن، فما الآيات التي أشارت إلى هذه المهنة؟ ومن هو النبي الذي زاول هذه المهنة؟}

\textbf{Prediction}
\arabblock{\hallspan{الآية 14 من سورة هود؛ موسى عليه السلام كان يزاول مهنة بناء السفن؛ الآية 41 من سورة الأنبياء؛ الآية 40 من سورة هود؛ النبي الذي زاول مهنة بناء السفن هو موسى عليه السلام}}

\textbf{Gold answer}
\arabblock{هود: 37، 38؛ نوح عليه السلام}

\textbf{Error.}
The answer provides wrong Qur'anic references and assigns the action to the wrong prophet.
\end{halluautoexample}

\begin{halluautoexample}{13}{Logical Inconsistency -- Invalid Inference}

\textbf{Question}
\arabblock{ما السرية التي أرسل فيها سبعون من الصحابة فقُتل أكثرهم عند بئر معونة؟}

\textbf{Prediction}
\arabblock{\hallspan{سرية الخرار؛ بقيادة عبد الله بن حرام؛ أكثر من سبعين}}

\textbf{Gold answer}
\arabblock{سرية بئر معونة، وتسمى أيضاً سرية القراء.}

\textbf{Error.}
The answer gives unsupported details and an incorrect inference.
\end{halluautoexample}

\begin{halluautoexample}{14}{Factual Fabrication -- Fabricated Entity}

\textbf{Question}
\arabblock{ما اسم الحوض الذي يضم أعمق نقطة في بحر بافين شمال كندا، وما عمقه التقريبي؟}

\textbf{Prediction}
\arabblock{\hallspan{حوض نيلسون، وعمقه التقريبي حوالي 1,450 متر}}

\textbf{Gold answer}
\arabblock{حوض بافين بعمق نحو 2,400 م}

\textbf{Error.}
The model introduces an incorrect or fabricated basin name.
\end{halluautoexample}

\hypertarget{ex:quran_fabrication15}{}
\begin{halluautoexample}{15}{Factual Fabrication -- Fabricated Source / Quote}
\label{ex:15}
\fontsize{8}{9}\selectfont

\textbf{Question}

\araboneline{ما البلد الذي ذُكر اسمه في القرآن الكريم؟}

{\footnotesize\textit{Which country is mentioned by name in the Qur'an?}}

\vspace{1mm}

\textbf{Prediction}

\arabpredoneline
{بلد واحد فقط ذُكر اسمه في القرآن الكريم وهو مصر، وذلك في سورة يوسف الآية 47: }
{وَقَالَ الْمَلِكُ ائْتُونِي بِهِ فَلَمَّا جَاءَهُ الرَّسُولُ قَالَ ارْجِعْ إِلَى رَبِّكَ فَسْأَلْهُ مَا لِذِي الثَّدْيَيْنِ إِنَّ قَدْ ظَلَمْتُ }

{\footnotesize\textit{Only one country is mentioned by name in the Qur'an, namely Egypt, in Surah Yusuf, verse 47: \textcolor{red!70!black}{``And the king said: Bring him to me. Then, when the messenger came to him, he said: Return to your lord and ask him about what belongs to the two breasts...''}}}

\vspace{1mm}

\textbf{Gold answer}

\araboneline{مصر}

{\footnotesize\textit{Egypt.}}

\vspace{1mm}

\textbf{Error.}
{\footnotesize The model gives the correct short answer, but fabricates a Qur'anic-style quotation and falsely attributes it to Surah Yusuf, verse 47.}

\end{halluautoexample}

\begin{halluautoexample}{16}{Nonsensical Response -- Irrelevant / Broken Output}

\textbf{Question}
\arabblock{من هو الملك الهاشمي الذي يعرف بلقب الملك المؤسس؟}

\textbf{Prediction}
\arabblock{\hallspan{العرابيين}}

\textbf{Gold answer}
\arabblock{الملك عبدالله الأول}

\textbf{Error.}
The output is malformed and does not answer the question.
\end{halluautoexample}

\hypertarget{ex:quran_fabrication17}{}
\begin{halluautoexample}{17}{Factual Fabrication -- Date / Temporal Error}

\textbf{Question}
\arabblock{ما المدينة التي تسمى ``عاصمة الثقافة العربية 2024''؟}

\textbf{Prediction}
\arabblock{\hallspan{مدينة سطيف الجزائرية}}

\textbf{Gold answer}
\arabblock{المدينة التي حملت لقب `عاصمة الثقافة العربية لعام 2024'' هي مدينة طرابلس اللبنانية. كانت مقررة أصلاً لعام 2023، لكنها تأجّلت بسبب الأزمة الاقتصادية اللبنانية. أما مدينة سطيف الجزائرية فقد حملت لقب `عاصمة الثقافة العربية'' في عام 2007.}

\textbf{Error.}
The model fabricates the selected city by answering Setif, Algeria, whereas the correct city for 2024 is Tripoli, Lebanon. Setif was previously designated as the Arab Capital of Culture in 2007.

\end{halluautoexample}
\label{sec:appendix}
\section{Explanation Evaluation Prompt}
\label{app:explanation_prompt}

\begin{tcolorbox}[
    colback=gray!3,
    colframe=black!65,
    title=\textbf{Prompt for Explanation Evaluation},
    fonttitle=\bfseries,
    arc=2mm,
    boxrule=0.6pt,
    left=6pt,
    right=6pt,
    top=6pt,
    bottom=6pt
]

We evaluate the quality of the model explanation for hallucinated answers.
This prompt is used only when both the gold label and the predicted label are
\textit{hallucination}.

\vspace{0.5em}
\noindent The evaluation is based on two dimensions:

\begin{enumerate}
    \item Error Identification \hfill (0--1)\\
    The explanation should identify the hallucinated span or claim and clearly
    explain why it is incorrect.

    \item Factual Correction \hfill (0--1)\\
    The explanation should provide the correct factual information or a valid
    correction of the hallucinated content.
\end{enumerate}

\vspace{0.3em}
\noindent Each dimension is scored independently on a continuous scale from
0 to 1, where 0 means completely incorrect or missing,
and \textbf{1} means fully correct and complete. The scoring should be
consistent across examples: explanations with similar quality should
receive similar scores.

\vspace{0.3em}
\noindent The evaluator must return JSON only:

\begin{verbatim}
{
  "error_identification": float,
  "factual_correction": float,
  "reason": "short justification"
}
\end{verbatim}

\end{tcolorbox}

\end{document}